%% file: main.tex
\def\year{2022}\relax
\documentclass[letterpaper]{article} 
\usepackage{aaai21}  
\usepackage{times}  
\usepackage{helvet} 
\usepackage{courier}  
\usepackage[hyphens]{url}  

\usepackage{ulem}

\usepackage{multirow}
\usepackage{graphicx} 
\usepackage{natbib}  
\usepackage{caption} 
\usepackage{times}
\usepackage{latexsym}

\usepackage{comment}

\usepackage{changepage}
\usepackage{graphicx}
\usepackage{pgfplots}
\usepgfplotslibrary{groupplots}
\usepackage{textcomp}
\usepackage{todonotes}
\usepackage{soul}
\usepackage{amsmath}

\sethlcolor{lightgray}

\usepackage{caption}
\usepackage{subcaption}
\usepackage[english]{babel}

\title{Catch Me If You Can: Deceiving Stance Detection and Geotagging Models to Protect Privacy of Individuals on Twitter}
\author {
    Dilara Dogan,\textsuperscript{\rm 1}
    Bahadir Altun,\textsuperscript{\rm 1}
    Muhammed Said Zengin,\textsuperscript{\rm 1} 
    Mucahid Kutlu,\textsuperscript{\rm 1} 
   {\normalfont and}
    Tamer Elsayed\textsuperscript{\rm 2}  \\
}
\affiliations {
   
    \textsuperscript{\rm 1} TOBB University of Economics and Technology
    \\
    \textsuperscript{\rm 2} Qatar University \\
    dilara.dogan@etu.edu.tr, i.altun@etu.edu.tr,  muhammedsaid.zengin@etu.edu.tr,m.kutlu@etu.edu.tr,  telsayed@qu.edu.qa \\
}

\begin{document}
\maketitle

\begin{abstract}
The recent advances in natural language processing have yielded many exciting developments in text analysis and language understanding models; however, these models can also be used to track people, bringing severe privacy concerns. In this work, we investigate what individuals can do to avoid being detected by those models while using social media platforms. We ground our investigation in two exposure-risky tasks, stance detection and geotagging. We explore a variety of simple techniques for modifying text, such as inserting typos in salient words, paraphrasing, and adding dummy social media posts. Our experiments show that the performance of BERT-based models fined tuned for stance detection decreases significantly due to typos, but it is not affected by paraphrasing. Moreover, we find that typos have minimal impact on state-of-the-art geotagging models due to their increased reliance on social networks; however, we show that users can deceive those models by interacting with different users, reducing their performance by almost 50\%.

\end{abstract}

\section{Introduction}

Recent developments in artificial intelligence (AI), especially in  natural language processing (NLP), bring many opportunities to deploy AI models in real life, such as human-like speaking personal assistants and accurate machine translation models. 
The massive amount of data available over social media platforms 
enables the development of increasingly-accurate models that predict lots of ``implicit'' information about users, e.g., location~\cite{rahimi2018semi}, stance on various issues~\cite{kuccuk2020stance}, age, gender~\cite{morgan2017predicting}, mental health~\cite{sekulic2019adapting}, race, or ethnicity~\cite{preoctiuc2018user} among others.

Several of those developed models can be used for a good cause. 
For instance, we can utilize stance detection models for fact-checking  \cite{baly2018integrating} and social polarization analysis \cite{rashed2020embeddings}. 
 Similarly, geotagging models can be used to identify affected areas during natural disasters~\cite{ghahremanlou2015geotagging}, and reduce bias in data collected for public opinion prediction~\cite{dwi2015twitter}.

Despite their beneficial use cases, many people have a legitimate privacy concern, as those platforms have access to too much information about people. While we also share similar concerns about using private information of people for commercial activities such as targeted ads, we believe that there is a more severe problem: \textit{the data is accessible by anyone, not only by the social media companies}.  

Understandably, many people opt for having public profiles instead of protected ones. Therefore, any person can crawl massive amounts of data these platforms provide and develop AI models for various reasons, including unethical ones such as 
surveillance of individuals. 
For instance, 
automatic stance detection methods might be problematic for people living in countries with limited freedom of speech or in a highly polarized society. Similarly, geotagging methods can be used to identify and expose where a particular individual lives, which might even  cause physical harm to individuals. Therefore, these models can be easily weaponized if used by the wrong people. Furthermore, recent developments in NLP, e.g., transformer models like BERT \cite{devlin2019bert} and GPT-3 \cite{brown2020language}, enabled the development of  effective NLP models with a minimal effort, requiring only a few lines of code and a labeled dataset. Therefore, the data can be used even by people with limited NLP knowledge and coding skills.

Due to  privacy concerns, many social media users tend to hide their identity behind nicknames or their location behind anonymous terms, e.g., ``earth''.
However, too much information is still subject to be revealed (hence exposing their owners) due to the success of modern models in leveraging unintentionally-available or implicit clues in their social media traces.  
\textit{We believe that individuals should be able to opt for not being tracked by those AI models, if they so desire.} However, exploiting the available data over social media platforms leaves individuals vulnerable.

In this work, we investigate how users can protect their privacy against AI models by themselves while using social media platforms. We focus on two ``exposure-risky'' tasks,
stance detection and geotagging,
 because individuals' physical location and  stance on various issues can potentially be used against them as mentioned above. We identify state-of-the-art models for each task and explore how to fool them using simple text manipulation techniques, such as inserting strategic typos, paraphrasing, and adding extra text 
  to provide  a list of recommendations to reduce the likelihood of being detected by AI models.


In particular, we seek answers to the following research questions. \textbf{RQ1:} \textit{What are the most effective text manipulation methods which fool state-of-the-art stance detection models without changing the semantics of the text?}
We found that BERT-based models are vulnerable to typos; their performance decays significantly for stance detection when typos are introduced in the social posts by additional spaces or changing/shuffling their letters. However, we were not able to fool BERT-based models by paraphrasing. 

 \textbf{RQ2:} \textit{What are the most effective methods to fool state-of-the-art geotagging models?}
 Our experiments show that state-of-the-art  geotagging models are slightly affected by adding typos and mentioning various city names due to the models' reliance on social networks in addition to the posts' content. However, we found that interaction with the social network by posting dummy tweets mentioning various users is effective in fooling geotagging models.  

\textbf{RQ3:} \textit{Which method to fool models has the least side effects regarding readability and changing semantics?} 
In our  analysis, 
we found that shuffling word letters can cause unreadable tweets. In addition, carelessly removing hashtags might change the semantics of tweets. Furthermore, inserting space characters  rarely changes semantics and makes tweets unreadable. 
However, the other methods we apply automatically do not cause any semantic change or not-readable tweets.

Our contribution in this work is three-fold. 
(1) While there exist studies exploring how to fool AI models, we address the problem from a different perspective. We investigate what a random social media user, who might have no idea about how those AI models work, can do by himself/herself to protect his/her privacy. 
(2) We investigate the impact of 15 different methods to fool stance detection and geotagging models, 
providing recommendations accordingly.
(3) We release our code and data to support the reproducibility of our experiments.\footnote{The url is hidden due to blind-review process.}

\section{Related Work}\label{sec_related}

A number of researchers investigated adversarial attacks and defense mechanisms for various tasks \cite{ren2020adversarial}. Our work can also be considered an investigation of adversarial attacks against NLP models. However, we have a thoroughly opposite perspective, such that social media users are not ``attackers'' because we believe that they are potential victims and explore how they can “defend” their privacy. Ignoring different perspectives on this issue, we now compare our study against prior work on adversarial attacks, especially attacks for NLP models. 

\citet{chen2020badnl} investigated \textit{backdoor attacks} in which training data of models are manipulated such that targeted NLP models fail when specific triggers (e.g., words) are used, but work as usual  with clean data. 
\citet{yang2021careful} show that 
  changing only a single word embedding vector is an effective method to hack sentiment analysis and sentence-pair classification models without causing any deterioration in the results of the existing clean samples. \citet{dai2019backdoor} demonstrate that a backdoor attack by inserting trigger sentences into  training data of an LSTM-based sentiment analysis model is highly effective. 
\citet{kurita2020weight} 
compare 
various backdoor attacks 
mentioned by \citet{gu2017badnets} for sentiment analysis, toxicity detection, and spam detection tasks. 
They show that 
attack successes change for each NLP task. 
\citet{sun2020natural} 
introduces \textit{natural backdoor attacks} that are 
hard to be  
noticed by humans  and grammar correction systems. 
Sun shows that the natural backdoor attacks are highly successful for text classification problems. In our work, we assume a black-box model such that we do not have  access to train data, and we aim to fool already trained models. 
However, backdoor attacks assume that they can  affect the training phase of AI models.

A number of researchers also explored vulnerabilities of NLP models in a black-box setting by changing the test data. 
The methods prior work investigated can be grouped into three categories: 1) character-level changes in which words are written with various forms of spelling errors, 2) word-level changes in which words are replaced, removed, or added, and 3) sentence-level changes in which new sentences or phrases are added or existing ones removed or paraphrased.  \textbf{Table \ref{tab_prior_work}} shows these adversarial attack methods investigated by prior work.

\begin{table*}[htb]
\tiny
\begin{center}
\begin{tabular}{| p{0.05cm} |p{4cm}| p{5.5cm} | p{6.5cm} |}
\hline
& \textbf{Method} & \textbf{Example} & \textbf{Study}  \\ \hline
\multirow{ 10}{*}{\rotatebox{90}{Character Level}} & Insertion  & apple $\rightarrow$ applee & \cite{sun2020adv} \\ \cline{2-4} 
& Deletion & school $\rightarrow$ schol & \cite{sun2020adv,li2019textbugger} \\ \cline{2-4} 
&Character swap & hello $\rightarrow$ hlelo & \cite{sun2020adv,li2019textbugger} \\ \cline{2-4}  
&Using different words pronounced same/similar & egg $\rightarrow$ agg & \cite{sun2020adv}  \\ \cline{2-4} 
&Replacing characters with the nearby ones in a keyboard & shy $\rightarrow$ why & \cite{schiller2021stance,sun2020adv,srivastava2020noisy,li2019textbugger,belinkov2018synthetic}  \\ \cline{2-4}  
&Replacing letters w/ visually similar characters & foolish  $\rightarrow$ fo0lish & \cite{dai2019backdoor,li2019textbugger,morris2020reevaluating,liang2018deep}  \\ \cline{2-4}  
&Inserting a space within a word & school $\rightarrow$  sc hool & \cite{sun2020adv,li2019textbugger}  \\ \cline{2-4} 
&Mistyping any character & talk $\rightarrow$ taln & \cite{sun2020adv,ebrahimi2018hotflip}  \\ \cline{2-4} 
&Common misspelling & film  $\rightarrow$ flim  & \cite{liang2018deep}  \\ \cline{2-4} 
&Middle Character shuffle &noise$\rightarrow$ nisoe& \cite{belinkov2018synthetic}  \\ \hline \hline
\multirow{ 5}{*}{\rotatebox{90}{Word Level}}&Replace words with semantically similar ones & awful $\rightarrow$ terribly & \cite{jin2020bert,li2019textbugger,niu2018adversarial,ebrahimi2018hotflip} \\ \cline{2-4} 
&Swap adjacent words & ``I don’t want you to go" $\rightarrow$ ``I don’t want to you go" & \cite{niu2018adversarial} \\\cline{2-4} 
&Remove Stopwords & Ben ate \sout{the} carrot & \cite{niu2018adversarial} \\\cline{2-4} 
&Insert a word & The Uganda Securities Exchange (USE) is the \textcolor{red}{\textit{historic}} principal stock
exchange of Uganda. & \cite{liang2018deep}\\\cline{2-4} 
&Remove a word & The Old Harbor Reservation Parkways are three \sout{historic} roads in the
Old Harbor area of Boston.  & \cite{liang2018deep}\\\hline \hline

\multirow{ 5}{*}{\rotatebox{90}{Sentence Level\ \ \ \ \ }} &Add a sentence & he Old Harbor Reservation Parkways are three historic roads in the
Old Harbor area of Boston. \textcolor{red}{\textit{Some exhibitions of Navy aircrafts were
held here.}}   &  \cite{jia2017adversarial,liang2018deep} \\\cline{2-4} 
&Paraphrase a sentence & “How old are you” $\rightarrow$ “What’s your age” & \cite{niu2018adversarial}  \\\cline{2-4} 
&Paraphrasing a phrase & the actual composer is \sout{different from} \textcolor{red}{\textit{not}}  the artist& \cite{liang2018deep} \\\cline{2-4} 
&Removing a phrase & \sout{promotion of world security}, improvement of economic conditions & \cite{liang2018deep} \\\cline{2-4} 
&Grammar Errors & “He \sout{doesn’t} \textcolor{red}{\textit{don't}} like cakes”   & \cite{niu2018adversarial}\\\hline

\end{tabular}
\caption{Adversarial attacks used in prior work. Red and italics words represent the added words. }
\label{tab_prior_work}
\end{center}
\end{table*}

Among the methods used by prior work, we also use  middle character shuffle \cite{belinkov2018synthetic} and inserting a space character \cite{sun2020adv,li2019textbugger} methods. In addition, some of these methods can be considered  similar to ours. For instance,  \citet{dai2019backdoor,li2019textbugger,morris2020reevaluating}, and \citet{liang2018deep} replace some letters with visually similar ones, but we use a different replacement scheme. 

\citet{liang2018deep} detect the most frequent phrases in the respective training dataset and remove/add them to fool NLP models, assuming they have access to the training data. However, we remove/add hashtags without any analysis of the train data. \citet{jin2020bert}, \citet{li2019textbugger},  and \citet{ebrahimi2018hotflip} replace words with semantically similar ones using word embeddings.  We replace words with their synonyms or use uncommon names of famous people. 

\citet{niu2018adversarial} paraphrase sentences using Pointer-Generator Networks. \citet{liang2018deep} paraphrase phrases using the approach of \citet{barzilay2001extracting}.   We also use various  paraphrasing methods, such as using  idioms. However, our paraphrasing methods focus on fooling methods instead of just expressing statements in a different way. 

\citet{schiller2021stance} investigate the robustness of stance detection models using three different adversarial attacks which are 1) adding the tautology “and false is not true” at the beginning of each sentence, 2) introducing spelling errors by character swaps and substitutions, and 3) paraphrasing by back-translation. They report that transformer-based models have serious robustness problem due to the overfitting on biases of training data.  They  specifically focus on  the stance detection task, but our work covers exposure-risky tasks including  stance detection and geo-tagging tasks. While their methods to fool the models can be considered  similar to some of our methods,   we use  additional fooling methods such as using idioms.

\citet{jia2017adversarial} add manually selected sentences to create adversarial examples for reading comprehension systems. Our mentioning a particular city method to fool geotagging models can be considered a similar approach.

To our knowledge, some of our methods have not been used by prior work, including interaction with other users, removing spaces, and using idioms. In addition, our targeted tasks are different from prior work. In particular, prior work investigated adversary attacks for various NLP tasks, including sentiment analysis \cite{jin2020bert,dai2019backdoor,li2019textbugger}, question answering \cite{jia2017adversarial}, dialogue generation \cite{niu2018adversarial}, machine translation \cite{belinkov2018synthetic},  toxicity detection \cite{kurita2020weight}, textual entailment \cite{jin2020bert}, and spam detection \cite{kurita2020weight}. However, to our knowledge, this is the first study focusing on the geotagging task. Furthermore, the goal of our study is to identify methods that will help individuals hide their personal information from AI 
models on social media. Therefore, all our methods can be applied manually. We leave developing a tool to protect privacy as future work. On the other hand, prior work focused on increasing the robustness of NLP models by exploring their vulnerabilities \cite{liang2018deep}, generating adversary examples for adversary training \cite{li2019textbugger,jin2020bert}, and enhancing models' architectures for noisy data \cite{muller2019enhancing, niu2018adversarial}.

Our work is also related to studies in ethics in NLP. \citet{mieskes2017quantitative} investigated 
ethical issues in data collection and sharing  in NLP. She suggests  
anonymizing  users instead of directly using their sensitive data. 
However,
%
\citet{feyisetan2020workshop} 
state that using  anonymized data 
does not actually solve the privacy problem. In addition, 
\citet{silva2020using} show that NLP Tools such as NLTK\footnote{https://www.nltk.org/}, Stanford CoreNLP\footnote{https://stanfordnlp.github.io/CoreNLP/}, and SpaCy\footnote{https://spacy.io/} can tag personal information on anonymized data. 
In our work, we focus on how to protect privacy while using social media platforms 

\section{Targeted Tasks and Models}\label{sec_targeted_tasks}

We focus on two  tasks, namely stance detection and geotagging, which might cause privacy concerns if accurately detected. Here we define the tasks and describe the state-of-the-art work we used in our study.

\subsection{Stance Detection}

Stance detection is the task of determining whether the author of a given text is favoring, against, or neutral towards a target or proposition~\cite{mohammad2016semeval}.  
\citet{ghosh2019stance} compare many stance detection models and report that fine-tuned BERT model yields the best prediction performance on the popular SemEval 2016 Task 6A dataset~\cite{mohammad2016semeval}. Therefore, we use the fine-tuned BERT model as one of our models. However, BERT is pre-trained on clean text with few typos, while some of our methods insert typos.  Therefore, as an additional model, we use a fine-tuned  Twitter-RoBERTa\footnote{https://huggingface.co/cardiffnlp/twitter-roberta-base}, which is pre-trained with 58M tweets, thereby, might be more robust for typos. 

\begin{table*}[!htb]
\begin{center}
\tiny
\begin{tabular}{|p{0.05cm}|p{2cm}|p{6.2cm}|p{6.2cm}|}
\hline
& \textbf{Method} & \textbf{Original Tweet} & \textbf{Modified Tweet} \\ \hline 
\multirow{ 5}{*}{\rotatebox{90}{\textbf{Typos}\ \ \ \ \ \ \ \ \ \ \ \ \ \ \ \ \ \ \ \ \ }} &Remove Spaces & also what's up with this ridiculous weather ? ? it was raining this morning and now it's like super hot ! \#weather problems \#lame & also what's up with this \textbf{ridiculousweather} ? ? it \textbf{wasraining} this morning and now it's \textbf{likesuper} hot ! \#weather problems \#lame \\ \cline{2-4}
&Add Spaces & breaking 911 probably she made a promise to support gun rights to one citizen , while promising to ban guns to the other & \textbf{b reaking} 911 pro bably she made a promise to \textbf{su pport g un} rights to one \textbf{cit izen} , while \textbf{pro mising} to \textbf{b an} guns to the other \\ \cline{2-4}
&Shuffle Word Letters & adam smith usa because clearly hillary clinton is a champion for us all & adam smtih usa because \textbf{clarely hlliary clitonn} is a \textbf{champoin} for us all \\ \cline{2-4}
&Change Character & men and women should have equal rights, we are all human & men \begin{bfseries}änd w0men\end{bfseries} should have \begin{bfseries}equä\textbar\end{bfseries}   \begin{bfseries}r!ghts\end{bfseries} , we are all \begin{bfseries}humän\end{bfseries}  \\ \cline{2-4}
&Add Hash Signs & hillary clinton hillary for nh hope to see her in not cool soon&hillary clinton hillary \textbf{\#for} nh \textbf{\#hope} to \textbf{\#see} her \textbf{\#in} not \textbf{\#cool} soon\\ \hline  \hline
\multirow{ 8}{*}{\rotatebox{90}{\textbf{Paraphrasing\ \ \ \ \ \ \ \ \ \ \ \ \ \ \ \ \ \ \ \ \ \ \ \ \ \ \ \ \ \ \ \ \ \ \ \ \ \ \ \ \ \ \ }}}&Use Uncommon Names & hillary clinton hillary for nh hope to see her in not cool soon & hillary \textbf{diane} clinton for us hope to see her in not cool soon \\ \cline{2-4}
&Use Antonyms Together & there's no more normal rains anymore always storms, heavy and flooding & contrary to \textbf{normal} there is more \textbf{abnormal} rain now always storms, heavy and floods \\ \cline{2-4}
&Add Hashtag & it's time that we move from good words to good works, from sound bites to sound solutions hillary clinton \#ready for hillary & it's time that we move from good words to good works, from sound bites to sound solutions hillary clinton \#ready for hillary \textbf{\#usa \#decisition \#time \#election \#future} \\ \cline{2-4}
&Remove Hashtags & \#fiona bruce wants a government that forces women to have children, and then refuses to financially help them \#body autonomy & bruce wants a government that forces women to have children , and then refuses to financially help them \\ \cline{2-4}
&Use Synonyms & generate belief in quality existence for everyone especially children in that community kitti ngt on 2016 & generate belief in quality existence for everyone especially \textbf{kids} in that community kitti ngt on 2016 \\ \cline{2-4}
&Use Idioms & also what's up with this ridiculous weather ? ? it was raining this morning and now it 's like super hot ! \#weatherproblems \#lame & also what's up with this ridiculous weather ? ? it was raining this morning and now \textbf{it 's dog days} ! \#weatherproblems \#lame \\ \cline{2-4}
&Remove  Words & success hillary clinton said she 's receiving a constant barrage of attacks from the right great job , guys keep it up ! & success hillary clinton said she 's receiving a constant barrage of attacks from the right great job \\ \cline{2-4}
&Use  Negations & the irish national school system is secular under law we can reaffirm secularism by going through the courts ! humanism ireland & the irish national school system is \textbf{not non secular} under law we can reaffirm secularism by going through the courts ! humanism ireland \\ \hline

\end{tabular}
\caption{Sample tweets along with the modified versions to deceive models. The modified words are boldfaced.}
\label{tab_samples}
\end{center}
\end{table*}

\subsection{Geotagging}

Research on geotagging can be broadly categorized into two groups: detecting the location of tweets~\cite{yavuz2016implicit} or users~\cite{rahimi2018semi}. In this study, we focus on the latter one. We use two state-of-the-art geotagging studies in our work. 
The first is based on Graph Convolutional Networks (\textbf{GCN})~\cite{rahimi2018semi}, which is a hybrid model that uses both textual content and user network in prediction, where  
text is represented as bag-of-words and graphs are generated from user mentions. 
The second is \textbf{MLP-TXT+NET}~\cite{rahimi2018semi}, which uses a multilayer perceptron in which each timeline is represented by concatenating bag-of-words vectors of tweets and user network. 


\section{Problem Definition}\label{sec_problem_def}

Our goal is to find ways that enable users to share their posts without risking the exposure of their private information by AI models. Therefore, we use methods that change the posts (or profiles) while keeping the same (or at least similar) semantics. Formally, let $t$ be a tweet posted by a user $u$, $f$ be our text or profile manipulation method, $m_s$ be a stance detection model, and $m_g$ be a geotagging model. An ideal method should have the following properties:
\begin{itemize}
    \item \textbf{Maintaining Semantics:} Semantics of $f(t)$ should be similar to the semantics of $t$. Similarly, $f(u)$ should have the same or similar tweets to $u$.
    \item \textbf{Minimal Side Effect:} To make methods applicable in real-life, the side effects of using them should be minimal. For instance, inserting intentional typos might effectively deceive AI models; however, tweets with typos look unprofessional and have low readability. 
    \item \textbf{Deceiving AI Models:} The modified tweets or user profiles should be able to deceive AI models, yielding inaccurate predictions. In particular, if $m_s(t)$ yields a correct stance, then $m_s(f(t))$ should yield a different one. 
    In addition, the accuracy of geotagging models is usually defined by the distance between predicted and actual locations; therefore, $m_g(f(u))$ should be farther from the actual location than $m_g(u)$.
\end{itemize}

\section{Methods to Deceive the Models}\label{sec_methods}

This section presents the methods we study to fool stance detection and geotagging models. The methods can be grouped into three groups including 1) inserting typos, 2) paraphrasing tweets, and 3) adding additional tweets to user profiles. As we focus on stance detection of tweets, we apply only inserting typos and paraphrasing tweets for the stance detection task. However, geotagging models predict locations based on user profiles. Therefore, in addition to methods modifying tweets' content, we also explore the impact of methods that add additional tweets to user profiles. 


In order to determine methods changing tweet contents, we conducted a manual text modification study and explored how we should post messages in order to hide our personal information.  In particular, 
we first fine-tuned the BERT model for stance detection using SemEval Task 6 dataset~\cite{mohammad2016semeval}. 
Subsequently, we randomly sampled tweets that the fine-tuned BERT model could predict their stance correctly. Subsequently, we changed the tweet contents manually to fool the model and determined our methods. 
We heuristically developed the methods that add additional tweets to user profiles.  

Now, we explain the methods used in this work. We consider
that tweets are modified manually for now. In the following section we discuss how we can automate some of these methods. 
\textbf{Table \ref{tab_samples}} provides a sample tweet for each of our methods.  

\subsection{Intentional Typos}\label{sec_typo}

BERT models generate embeddings for subwords based on its vocabulary. If it encounters an out-of-vocabulary word, it slices the word into subwords and generate an embedding for each. By inserting typos, our goal is to cause more out-of-vocabulary words and cause BERT and Roberta models to create embeddings for unrelated subwords. For instance, writing the word ``against" as ``aganist" would cause BERT to generate embeddings for ``ag", ``-ani", and ``-st" tokens instead a single embedding for the word ``against". 
Geotagging models we use in our study utilize bag-of-words representation for tweet contents. Therefore, by inserting typos, we can reduce the number of words used in bag-of-words representation because the words with typos are less likely to exist in the training datasets. Now we explain our methods for various typos. 

\noindent
\textbf{Remove Spaces:} Space characters are essential to separate words. This method removes some space characters, thus combining the respective words.  Removing all spaces would make the text unreadable; therefore, we select important words (that might yield accurate prediction) and combine them with the following or preceding word. We apply this as long as we think the tweet is readable. 

\noindent
\textbf{Add Spaces:} In this method, we add a space character within the letters of an important word. 

\noindent
 \textbf{Shuffle:} 
 Similar to the character swapping method used by prior work  \cite{sun2020adv,li2019textbugger}, we change the order of letters in selected words while keeping the first and last words intact, inspired by the urban legend known as ``Typoglycemia''\footnote{https://en.wikipedia.org/wiki/Typoglycemia}. While this method can actually result in unreadable texts in some cases\footnote{https://www.mrc-cbu.cam.ac.uk/people/matt.davis/cmabridge/}, we apply this method if words are still recognizable in our manual modifications. 
 
 \noindent
 \textbf{Change Characters:}  We use popular writing styles used in social media platforms in which some letters are replaced with others with a similar appearance or pronunciation. In particular, our replacement procedure is as follows: a $\rightarrow$ \"a, i$\rightarrow$ !, l$\rightarrow$ \textbar , o$\rightarrow$ 0,  ae$\rightarrow$ $\ae$, $to \rightarrow 2$, $for \rightarrow 4$, and $great \rightarrow gr8$. While the resultant tweets are readable, we note that they do not look professional, reducing their real-life applicability. 
 
 
\noindent
\textbf{Add Hash Signs:} Hashtags (HT) are generally used to indicate important topics. In this method, we add \# sign to words (i.e., making them hashtags) which are  
 \textit{unimportant} for stance detection. Therefore, the model might be distracted by giving more attention to unimportant words, possibly yielding inaccurate predictions. 
 

\subsection{Paraphrasing}\label{sec_para}
In this set of methods, our goal is to make significant changes to the tweet content while maintaining its semantics. The main intuition in these methods is to exploit the bias that models learn from the datasets. For instance, even though BERT models generate contextualized word embeddings, \citet{niven2019probing} report that BERT's predictions are affected by the presence of cue words, especially the word ``not''. Therefore, if a word appears a lot in the training data with a specific label, the presence of that word, even though it is not directly related to the label, can affect the outcome. 

One of the main challenges we encountered during our manual text modification study was that we tried to modify texts written by others. Furthermore, social media posts might include dialects, incomplete sentences, and improper use of language with many grammar mistakes. We did our best to have meaningful and regular English sentences; however, as our primary goal is to explore the impact of some specific expressions, in some cases, the resultant text might sound weird or have unusual use of language.  Now we explain these methods for the stance detection task. 

\noindent
\textbf{Use Uncommon Names:} Instead of mentioning the full name 
of certain people, we use either their names' abbreviations (e.g., ``HC'' for ``Hillary Clinton'') or longer forms (e.g., ``Hillary Diane Clinton''). 
 
\noindent \textbf{Use Antonyms Together:} Using antonym of a word reverses the meaning. Thus, using two antonyms (e.g., normal and abnormal) together might confuse models while maintaining the meaning. 

\noindent \textbf{Add Hashtag:} Hashtags might be beneficial to predict the stance of a given tweet. Therefore, in order to confuse models, we add hashtags that are ``neutral'' to the stance of the tweet, e.g., ``\#monday'' and ``\#future''. 

\noindent \textbf{Remove Hashtag:} In this method, we remove hashtags that will not spoil the meaning.

\noindent \textbf{Use Synonyms:}  We replace words with their synonyms whenever possible (e.g., $children \rightarrow kids$).   

\noindent \textbf{Use Idioms:} Semantic analysis of idioms remains  a challenging task for language models. Therefore, in this method, we use idioms whenever applicable, e.g., ``brass monkey'' for ``extremely cold weather'' or ``raining cats and dogs'' for ``heavy rain''. 

\noindent \textbf{Remove Words:} 
We remove some words of the tweet that do not directly affect the meaning. 

\noindent \textbf{Use Negations:}  Negations might confuse models because they reverse the meaning of the  words used together with the negation words (e.g., ``not'' and ``without''). Therefore, in this method, we 
replace a positive expression with a negation word  and the opposite of the positive expression (e.g., ``is religious'' $\rightarrow$ ``is not nonreligious''). 

\subsection{Additional Tweets to User Profiles}\label{sec_add}

\noindent
\textbf{Mention City:} We can talk about a city even though we do not live there. This might deceive models due to mentioning a particular city regularly. In this method, we add a predefined set of tweets in which a particular location is mentioned (e.g., 
``Hawaii is beautiful!'' and ``The most expensive houses are in Hawaii'').



\noindent
\textbf{Mention Users:} State-of-the-art geotagging models use both text and social network for prediction, as described earlier. Therefore, in this method, we add tweets, with dummy text, which mention other users, changing their social network graph. Of course, mentioning random users can be considered  spamming. In real life, people might get in touch with their friends or celebrities living in different locations or local entities (e.g., local news channels) to apply this method. 

\section{Experimental Evaluation}\label{sec_experiments}

\subsection{Experimental Setup}

\textbf{Datasets.} For the stance detection task, we use the dataset of SemEval 2016 Task-6 \cite{mohammad2016semeval}, which consists of five topics: Atheism (AT), Climate Change (CC), Feminism (FM), Hillary Clinton (HC), and Legalization of Abortion (LA). Each tweet is labeled as one of the three labels: Against (A), Favor (F), None (N). The label distribution of training and test data is shown in \textbf{Table \ref{tab_stance_detection}}.

For the geotagging task, we use the popular GEOTEXT \cite{eisenstein2010latent} dataset, which 
includes 9K users and 370K tweets. Each user has a varying number of tweets, and latitude and longitude information as their labels. 
We use the same train, validation, and  test sets  shared by the original dataset creators. The ratio of train, validation, and test sets are 60\%, 20\%, and 20\%, respectively.   




\begin{table}[!htb]
\begin{center}
\small
\begin{tabular}{|l| l | l | l | l | l | l | l|}
\hline
&
\multicolumn{3}{c|}{\textbf{Train}} &
\multicolumn{3}{c|}{\textbf{Test}}\\
\hline
\textbf{Topic}&F&A&N&F&A&N\\ \hline
\textbf{AT}&92&304&117&32&160&28 \\ \hline
\textbf{CC}&212&15&168&123&11&35 \\ \hline
\textbf{FM}&210&328&126&58&183&44 \\ \hline
\textbf{HC}&112&361&166&45&172&78 \\ \hline
\textbf{LA}&105&334&164&46&189&45 \\ \hline
\end{tabular}
\end{center}
\caption{Label Distribution in Stance Detection Dataset. F: Favor, A: Against, N: None}
\label{tab_stance_detection}
\end{table}
\noindent \textbf{Evaluation Metrics}. To measure the performance of stance detection models, we report macro-average $F_1$ score across favor, against, and none classes for each topic. 
For the geotagging task, we report mean error (i.e., the distance in miles to the actual location) as in prior work \cite{rahimi2018semi}. 

\noindent \textbf{Implementation.} 
For the stance detection task, we fine-tune large uncased pre-trained BERT model\footnote{https://huggingface.co/bert-large-uncased} and pre-trained Twitter-RoBERTa-base model\footnote{https://huggingface.co/cardiffnlp/Twitter-RoBERTa-base} using the train set of each topic. We use 11 epochs 
with a batch size of 16. 
We found out that oversampling the rare classes by two improves the performance of BERT 
for Climate Change and Hillary Clinton topics due to the imbalanced distribution of labels. Therefore, we apply oversampling for these topics. 
For the geotagging task, we have used the implementation of GCN and MLP-TXT+NET\footnote{https://github.com/afshinrahimi/geographconv} shared by \citet{rahimi2018semi}.



\subsection{Manual Modification for Stance Detection}\label{sec_maual_results}
In our experiments with manual modification, we
first utilized the tweets we used for developing our methods. However, as mentioned above, the stance of these tweets are predicted accurately by the BERT model. In order to have a better sample and understand whether our methods yield correct classifications for tweets which are  misclassified before our modifications, we randomly sampled additional tweets to  be manually modified. 
\textbf{Table \ref{tab_manual_sample}} shows the distribution of topics in our sample, and the corresponding accuracy of BERT and Twitter-RoBERTa models.

\begin{table}[!htb]
\begin{center}
\small
\begin{tabular}{|l| c | c | c|}
\hline
\textbf{Topic}& \textbf{Tweets} & \textbf{BERT} & \textbf{Twitter-RoBERTa} \\ \hline
\textbf{AT}& 21 & 0.952 & 0.905\\ \hline
\textbf{CC}& 11 & 0.909 & 0.727\\ \hline
\textbf{FM}& 16 &  0.813 & 0.688\\ \hline
\textbf{HC}& 19 & 0.947 & 0.737\\ \hline
\textbf{LA}& 15 & 0.600 & 0.867\\ \hline \hline
\textbf{Overall} & 82 & 0.854 &  0.793 \\ \hline
\end{tabular}
\end{center}
\caption{The number of tweets we manually modified and accuracy of fine-tuned BERT and Twitter-RoBERTa models when original tweets are used for prediction for each topic.}
\label{tab_manual_sample}
\end{table}

\begin{table*}[!htb]
\small
\begin{center}
\begin{tabular}{|p{0.05 cm}|l| r | r |r |r |r ||r |r|r|r|}
\hline
&\multirow{ 2}{*}{\textbf{Methods}} & \multirow{2}{*}{\textbf{\# Trials}} &  \multicolumn{4}{c||}{\textbf{BERT}} & \multicolumn{4}{c|}{\textbf{Twitter-RoBERTa}}   \\ \cline{4-11}
& & & $T \rightarrow T$& $F \rightarrow F$  & $T \rightarrow F$ & \textbf{$F \rightarrow T$} & $T \rightarrow T$& $F \rightarrow F$  & $T \rightarrow F$ & \textbf{$F \rightarrow T$}  \\ \hline  \hline
\multirow{ 5}{*}{\rotatebox{90}{\textbf{Typos}}}&Change Character & 126 &  48\% &  17\% &  33\% &  2\% &  48\% &  21\%  &  23\% &  7\%       \\ \cline{2-11}
&Add Spaces & 84 &  63\% &  7\% &  30\% &  0\% & 64\% &  10\% &  21\% &  5\%      \\ \cline{2-11}
&Shuffle  & 90 &  50\% &  17\% &  33\% &  0\% &  68\% &  10\% &  19\% &  3\% \\ \cline{2-11} 
&Remove Spaces &  48 &  69\% &  25\% &  6\% &  0\% &  75\% &  13\% &  13\% &  0\%         \\ \cline{2-11}
&Add Hash Signs & 90 &  80\% &  10\% &  10\% &  0\% &  74\% &  17\% &  6\% &  3\% \\ \hline \hline
\multirow{9}{*}{\rotatebox{90}{\textbf{Paraphrasing}}}&Remove Hashtag &  57 &  65\% &  11\% &  19\% &  5\% &  70\% &  16\% &  9\% &  5\%    \\ \cline{2-11}
&Use Synonyms &  81 &  78\% &  15\% &  7\% &  0\% &  72\% &  16\% &  10\% &  2\%   \\ \cline{2-11}
&Add Hashtag &   75 &  71\% &  16\% &  9\% & 4\% &  76\% &  24\% &  0\% &  0\%  \\ \cline{2-11}
&Use Antonyms Together & 9 & 100\% & 0\% & 0\% &  0\% & 89\% &  0\% &  11\% &  0\%     \\ \cline{2-11}
&Use Uncommon Names &  21 &  76\% &  0\% &  24\% &  0\% &  48\% &  24\% &  10\% &  19\%  \\ \cline{2-11}
&Use Idioms & 27 &  96\% &  0\% &  4\% &  0\% &  89\% &  0\% &  0\% &  11\%     \\ \cline{2-11}
&Remove  Words&  12 &  75\% &  25\% &  0\% &  0\% &  50\% & 50\% &  0\% & 0\%    \\ \cline{2-11}
&Use Negations &  18 &  22\% &  11\% &  0\% &  6\% &  61\% &  28\% &  6\% &  6\%
 \\ \hline
\end{tabular}
\caption{The impact of our manual text modifications on the performance of BERT and TwitterRoberta models.  T stands for True, and F stands for False. $T \rightarrow F$ shows the ratio of the cases  we could change the correct prediction of the corresponding model to a false prediction by using the respective text modification method. Similarly, $F \rightarrow T$ shows the number of cases where a false prediction is changed to a correct prediction. $F \rightarrow F$ and $T \rightarrow T$  show the number of cases that do not change the prediction at all. Each method has been applied by three people.}
\label{tab_results}
\end{center}
\end{table*}

Obviously, the results of manual text modifications  depend on the person who performs the modifications. In order to reduce this bias in our results, the modifications are conducted by multiple people. In particular, one of the authors of this paper first manually modified the tweets using the methods explained in the previous section. For each tweet, the author developed three modified versions, applying a different method for each. Subsequently, two other authors of this work manually modified the tweets by applying the methods that have been used by the first author. For instance, if the first author applied shuffling, adding hashtags, and changing characters for a particular tweet to create the three modified versions, the other authors also applied the same techniques for that tweet. While they are required to apply the same method for a particular case, they were free on \textit{how} to apply it. For instance, they can change characters of different words and come up with different hashtags. This allowed us to control the number of trials for each method while creating different versions of a tweet by applying the same method. Eventually, we developed 738 ($=82 \times 3 \times 3$) modified tweets. 

%
\textbf{Table \ref{tab_results}} shows the number of trials for each method and the ratio of changes in the outcome of BERT and Twitter-RoBERTa  models from a true/false prediction to a true/false one. 
The number of trials varies for each method. This is because some methods are applicable only for particular instances. For example, in order to apply the ``Use Idioms" method, there should be a specific phrase that can be expressed with an idiom. Due to the varying number of trials for each method, we report the ratio of each case with respect to the number of trials.

\begin{table*}[htb]
\small
\begin{center}
\begin{tabular}{|p{0.1cm}|l| l |l|l||l|l|l||l|l|l|l|}
\hline
&\multirow{ 2}{*}{\textbf{Methods}} &  \multicolumn{3}{c||}{$T \rightarrow T$}& \multicolumn{3}{c||}{$T \rightarrow F$}   & \multicolumn{3}{c|}{$F \rightarrow T$}   \\ \cline{3-11}
& & \textbf{P1} & \textbf{P2} & \textbf{P3} & \textbf{P1} & \textbf{P2} & \textbf{P3} & \textbf{P1} & \textbf{P2} & \textbf{P3} \\ \hline  \hline
\multirow{ 5}{*}{\rotatebox{90}{\textbf{Typos}}}&Change Character &  22&19&20&	12&15&14 &1&0&1	 \\ \cline{2-2}
&Add Spaces & 18&16&19&	8&10&7&	0&0&0 \\ \cline{2-2}
&Shuffle  & 15&12&18&	10&13&7&	0&0&0\\ \cline{2-2} 
&Remove Spaces &   11&12&10&	1&0&2&	0&0&0     \\ \cline{2-2}
&Add Hash Signs & 22&26&24&	5&1&3&	0&0&0  \\ \hline \hline
\multirow{9}{*}{\rotatebox{90}{\textbf{Paraphrasing}}}&Remove Hashtag &  12&13&12&	4&3&4&	1&1&1   \\ \cline{2-2}
&Use Synonyms & 20&21&22&	3&2&1&	0&0&0  \\ \cline{2-2}
&Add Hashtag & 18&16&19&	2&4&1&	2&1&0 \\ \cline{2-2}
&Use Antonyms Together & 3&3&3&	0&0&0&	0&0&0 \\ \cline{2-2}
&Use Uncommon Names &  6&4&6&	1&3&1&	0&0&0\\ \cline{2-2}
&Use Idioms &   9&9&8&	0&0&1&	0&0&0\\ \cline{2-2}
&Remove  Words&3&3&3&	0&0&0&	0&0&0 \\ \cline{2-2}
&Double Negations & 5&5&5&	0&0&0&	1&0&0  \\ \hline \hline
\multicolumn{2}{|c|}{\textbf{Total}} & 164 & 159& 169& 46 & 51 &41 & 5 & 2 & 2  \\ \hline
\end{tabular}
\caption{The impact of manual text modifications of each person involved in the experiment (represented as P1, P2, and P3) on the predictions of BERT model. We show the number of cases for each prediction change type. We omit the results for $F \rightarrow F$ for simplification. }
\label{tab_results_bert_per_person}
\end{center}
\end{table*}

Regarding \textbf{RQ1}, we notice that paraphrasing 
 tweets usually does not change the prediction, showing that both models are able to catch the semantics of tweets even though we use different words. One of the unexpected results we obtained is the failure of deceiving models using idioms; models were able to detect stance correctly based on other words in the tweets.  

We also observe that both models are vulnerable to typos, echoing the findings of \citet{sun2020adv} for BERT model. We are able to deceive both models in around one third of the cases when we change characters with visually similar ones, split important words by adding spaces, and shuffle the letters in the middle of words. We observe that removing spaces is less effective than other typo-based methods, especially for BERT model. This is because the BERT tokenizer is able to correctly tokenize two consecutive words written without any space for some cases (e.g., ``ridiculousweather") 
While Twitter-RoBERTa has lower performance than BERT model on the original tweets (See Table \ref{tab_manual_sample}), its performance is less affected by our modifications compared to the BERT model. We think that this is due to being pre-trained with (typically noisy) tweets, enabling it to handle typos more effectively. 

Hashtags appear to be important for the BERT model. We could change a correct prediction to a wrong one in 19\% of the cases by removing hashtags. However, it is noteworthy that in 5\% of the cases when we removed hashtags, wrong predictions of both models have been changed to a correct one. Adding neutral hashtags or converting some words into hashtags also cause inaccurate predictions in 9\% and 10\% of the BERT predictions, respectively. However, Twitter-RoBERTa seems to be more robust to hashtag changes than BERT, as its performance is not affected by adding hashtags and is slightly affected by adding hash signs or removing hashtags. Note that these methods are applicable in platforms using hashtags. However, it is a highly popular writing style used in main social media platforms, including Twitter, Instagram, and Facebook. 

For people who do not want to be tracked due to their stances on various issues, changing the predicted stance to neutral is more important than changing it to an opposite stance. None of the tweets we manually changed has neutral label; however, when we use the original tweets for prediction, the number of tweets predicted as neutral is six and zero for BERT and Twitter-RoBERTa, respectively. When we use our modified tweets, BERT and Twitter-RoBERTa predict as neutral for 114 and 99 cases (out of 738), respectively, suggesting that modified versions are somewhat effective to change predictions to neutral ones. 


As manual modifications are biased to the people who modify the tweets, we investigate whether the success of methods change across them. \textbf{Table \ref{tab_results_bert_per_person}} shows the number of prediction changes of BERT for each method and each person modified the tweets. We omit the results for Twitter-RoBERTa for simplifying the discussion; however, we observe that Twitter-RoBERTa yields more stable results across people than BERT. In general, we observe that the performance of methods are similar across people, not changing any of our conclusions about comparison of methods. However, we also observe that P1, P2, and P3 could change correct predictions to false ones in 46, 51, and 41 cases, suggesting that it is also important how methods are applied. In general, paraphrasing methods have more stable results across people than typo-based methods. For instance, in ``Remove Words'' and ``Using antonyms together'' methods, all modifiers have exactly the same performance. This might be due to the limited flexibility of those methods. 

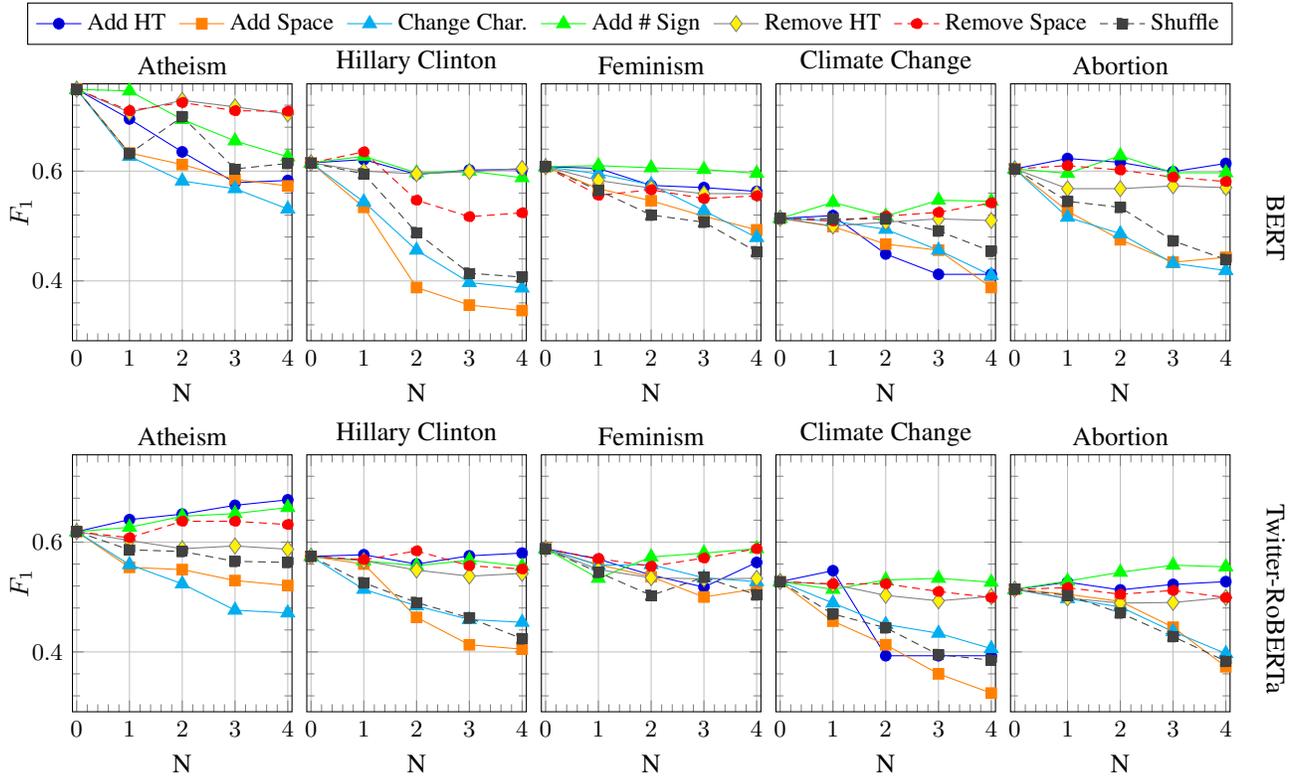
\begin{figure*}[!htb]
    
	   \input{stance_detection_all}
	   \input{roberta}
	
    \caption{Performance of fine-tuned  BERT and Twitter-RoBERTa models in stance detection task on the test data of the respective dataset of SemEval2016 for varying distortion numbers. For instance, $N=4$ means that the respective method has been applied for four words. The upper row shows $F_1$ score of BERT and the bottom row shows the $F_1$ score of Twitter-RoBERTa.}
    \label{figure_results_all}
\end{figure*}


\begin{figure*}[!htb]
    \centering
    \input{readability_and_semantic}
    \caption{Readability and semantic change analysis in stance detection task for varying number of distorted words. In particular, we manually analyzed 140 cases to understand whether the resultant tweets when our methods are applied automatically are readable and have the same semantic with the original tweets. y-axis show the ratio of readable tweets and the ratio of tweets without any semantic change among tweets we manually inspected. x-axis represent the number of words affected by our methods.}
    \label{figure_readability_all}
\end{figure*}
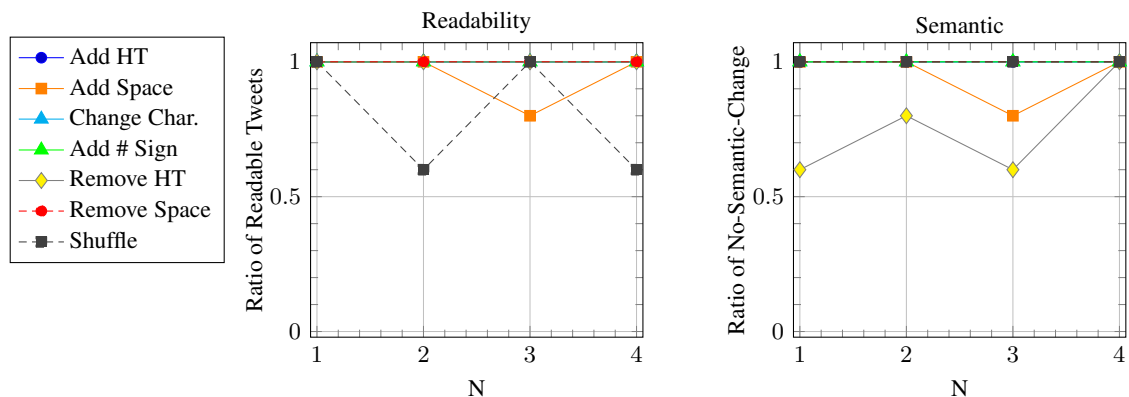

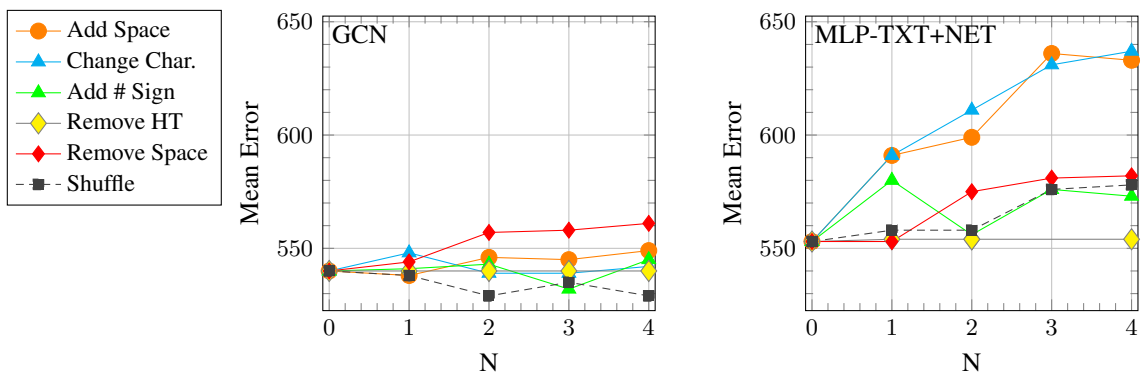
\begin{figure*}[!htb]
    \centering
    \input{geotagging_gcn}
    \caption{Mean error scores of geo-tagging models when our methods are applied for a varying number of times in all tweets of users in the test data of GEOTEXT dataset.}
    \label{figure_results_geo_all}
\end{figure*}

\subsection{Automatic Modification for Stance Detection} \label{sec_automatic_results}
In the previous experiment, we modified a subset of the tweets manually. Now, we investigate the impact of our methods when they are applied automatically on the whole dataset. It is challenging to paraphrase tweets based on our methods automatically.  Furthermore, they are not  effective in deceiving the  models. Therefore, we focus on methods that are potentially effective and can easily be applied automatically. In particular, in this set of experiments, we use ``Add Hash Signs'', ``Add Hashtag'', ``Remove Hashtag'', ``Change Character'', ``Shuffle Word Letters'', ``Add Spaces'', and ``Remove Space'' methods. In the manual modification, we did not put any restriction on how many words should be changed. Additionally, we manually selected ``important'' words to modify. For automatic modification, we introduce the parameter $N$ to denote the number of words to be modified and the number of hashtags to be removed/added. We vary $N$ from $0$ to $4$. 

In order to select the  words to be modified, we rank all words in a tweet based on their cosine similarity to the respective topic words (e.g., abortion) using fastText word embeddings~\cite{bojanowski2017enriching}, then we pick the closest $N$ words to modify. However, applying ``Remove Space''  method in consecutive words might cause unreadable tweets. Therefore, we select $N$ non-consecutive words for this method. 
Similarly, in order to detect 
words to be converted to hashtags in ``Add Hash Sign" method, we use the most distant ones to the topic words, because we convert the unimportant ones as explained in the previous section. In ``Add Hashtag'' method, we manually  defined a hashtag list for each topic and used $N$ of them. For instance, the hashtags for the abortion topic are \#MondayMotivation, \#goals, \#opinion, and \#thoughts. In addition, in  ``Shuffle'' method, 
if the selected word has 7 or more letters, we only change the position of the $2^{nd}$, $3^{rd}$, $4^{th}$, and $5^{th}$ letters, keeping the distance between the original position of a letter and its position in the jumbled version  at most three. This is 
because it is likely that words will not be readable when the position of a letter is changed a lot. 

In this experiment, we modify each tweet in the test set as explained above and report the performance of  BERT and Twitter-RoBERTa 
model fined tuned on the original train data. The results are shown in \textbf{Figure \ref{figure_results_all}}. 
Our observations regarding the results are as follows. Firstly, ``Remove Hashtag" seems the least effective method among others. On the other hand, similar to our experiments with  manually modified texts,  ``Change Character'', ``Add Space'', and ``Shuffle'' methods seem to be the most effective ones, decreasing the performance of BERT model by 28\%, 27\%, and 23\% on average across five topics when $N=4$, respectively. Similarly, ``Change Character'', ``Add Space'', and ``Shuffle'' decreases the performance of Twitter-RoBERTa by 20\%, 25\%, and 20\% on average across five topics when $N=4$, respectively.

Regarding ``Add Hashtag'' method, our findings are a mix. While it has a slight impact on both models' performance in most of the cases,  for the topic of atheism, it decreases the performance of BERT model, but improves Twitter-RoBERTa's performance. We observe a similar pattern for  ``Add Hash Sign'' method. 
This suggests that hashtags might have a correlation with the labels in the train data. Therefore,  it is risky to use hashtags without knowing the training data of models. 
  
Regarding BERT vs. Twitter-RoBERTa, BERT yields higher performance than Twitter-RoBERTa when $N=4$. Our first expectation was that  Twitter-RoBERTa would be less affected by the typos we introduced because it is pre-trained with noisy data. However,  there is no meaningful difference between models’ relative performance changes in our experiments with automatic modifications. 


In order to investigate whether the resultant texts after our automatic modifications are readable and have the same/similar meaning with the original tweet, we randomly sampled five tweets for each method and each $N$ value (i.e., $5 \times 7 \times 4 = 140$ cases in total) and manually checked whether they are readable and their semantics have changed. In our readability analysis, two authors of this paper manually checked each tweet and if a tweet contains at least one word which could not be identified by at least one of the authors, we  consider that tweet as  not readable. 
\textbf{Figure \ref{figure_readability_all}} shows the ratio of tweets that are readable and have the same/similar meaning as the original tweets among the tweets we inspected for each case. We see that all methods except ``Shuffle'' and ``Add Space'' yield readable tweets. ``Shuffle'' makes 40\% of tweets unreadable when $N=2$ and $N=4$, reducing its applicability. The following tweet is one of those unreadable ones where correct versions of words are written in parenthesis: ``\textit{ny investing big bkaenr (banker) bdus (buds) need to ratchet up their haillry (hillary) cares about the little polepe (people) propaganda}". 

Regarding the semantic change, we observe that none of our methods except ``Add Space" and ``Remove Hashtag" change the semantic of tweets. ``Remove Hashtag" causes semantic change because we observe that people use hashtags for important words in a tweet. For instance, in the following tweet, the removed hashtag (shown as strikethrough text) plays an essential role in the meaning of the sentence: ``\textit{agent 350 this is not a fantasy this is negligence collusion with criminal corporations acting with negligence to \st{\#ecocide}}". One might ask how ``Shuffle" does not cause any semantic change when tweets are not readable. For those cases, the changed words do not mean any other meaningful word. Therefore, we assumed that if someone can read them correctly (based on context), it would not cause any change in the meaning. Nevertheless, our qualitative analysis suggests that ``Shuffle'' and ``Remove Hashtag'' methods require special attention to keep semantics unchanged and tweets readable.   

\subsection{Automatic Modification for GeoTagging}\label{sec_automatic_geo}

Regarding \textbf{RQ-2}, we first apply the automatic methods used in the previous experiment for the geotagging task in this set of experiments. Similarly, we 
use parameter $N$ to control how many words are modified or hashtags are added/removed. We apply our methods to all tweets 
of users in the test set. As there is no topic in geotagging task, we randomly select words to be modified instead of using our fastText based similarity calculation. In our modifications, we do not change any mentioned user not to change social network used by the models. We do not use ``Add Hashtag'' method in geotagging, because there is no specific topic to be neutral. The results are shown in \textbf{Figure \ref{figure_results_geo_all}}.

  Generally, MLP-TXT+NET's performance decreases as $N$ increases in all methods, except the ``Remove Hashtag'' method. 
 In fact, ``Remove Hashtag'' method has no impact on the performance of both models. This might be because both models represent texts as bag-of-words and hashtags in the test set might not appear in the train set. 
 We observe that GCN's performance is slightly affected by our changes in the tweet content, suggesting that social network plays a critical role in its prediction.

\begin{figure}[!htb]
    \centering
    \input{additional_tweets}
    \caption{The impact of adding additional tweets created by our methods on the performance of two geo-tagging models, GCN and MLP-TXT+NET.}
    \label{figure_results_geo_add}
\end{figure}
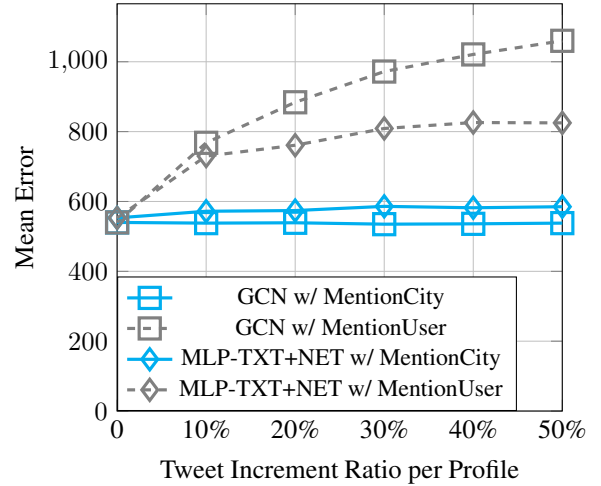

Next, we increase the number of tweets of each user 
using our ``Mention City" and ``Mention Users" methods, separately. We vary the increment ratio per profile from 10\% to 50\%. The results are shown in \textbf{Figure \ref{figure_results_geo_add}}. We observe that having many tweets mentioning cities has a limited impact on both models' performance. However, as we change the social network by mentioning random users, their performance decreases dramatically. Overall, our experiments suggest that users who want to hide their location from AI models can interact with users (e.g., celebrities and local entities in different places) located in various places, instead of adding typos or mentioning location names explicitly. 

\section{Ethical Discussion}
The main motivation of our work is to explore techniques that might be beneficial to reduce the negative impacts of AI models, which can be easily weaponized against humans. However, as weapons, the same AI models can be used to cause harm (i.e., surveillance of individuals) or to prevent harm (e.g., preventing the spread of misinformation and hateful messages). Therefore, people who spread misinformation or hate can also use similar techniques to get rid of AI models that might detect their toxic messages. On the other hand, our methods will also be helpful for random people who just do not want to be tracked by people they even do not know. Using the same analogy, our approaches can be considered armors if AI models are weapons. We believe that there should be available armors in the market, if we know that there are people with weapons. Our study makes a modest step towards this goal. We hope that our work will motivate other researchers to work on this important research direction and will find better ways than ours.

\section{Conclusion}\label{sec_conc}
In this work, we investigated how individuals can protect their privacy from AI models while using social media platforms. We focused on stance detection and geotagging tasks and explored fifteen different text-altering methods such as inserting typos into strategic words, paraphrasing, changing hashtags, and adding dummy social media posts.
Based on extensive experiments we conducted, our recommendations for people who do not want to be tracked by AI models on social media platforms are as follows. Firstly, paraphrasing methods do not work well to deceive the  models, suggesting that they successfully catch the semantics of texts. Although language models other than BERT might be utilized in real life, other large models will likely have similar performance at catching semantics. Secondly, changing characters with visually similar ones, splitting words by adding spaces, and shuffling the letter orders are effective in decreasing stance detection models' performance. However, these methods require special attention because the resultant texts might be unreadable.
Finally, in order to deceive geotagging models, the most effective way is to interact with a diverse set of users. 

In the future, we plan to extend our work in several directions. Firstly, we will explore other tasks focusing on predicting personal information about individuals such as race, ethnicity, and mental health. We also plan to develop more sophisticated methods to fool AI models. 
In addition, we plan to conduct a user study to investigate whether people are aware of AI models and their capabilities in detecting personal information. Furthermore, we will explore other datasets based on social media platforms other than Twitter to reduce platform-specific bias in our experiments. Moreover, we plan to reach vulnerable communities such as immigrants and extend our work based on their needs. Lastly, we will develop a tool which will modify  messages  to prevent tracking. We plan to leave the development of a tool as our final goal because a tool which does not work well might be harmful because of giving false hopes to people who would like to use it. Therefore, we will also explore explainable artificial intelligence techniques so that the users will be able to interpret its output and act accordingly.


\section*{Acknowledgment} This study was funded by the Scientific and Technological Research Council of Turkey (TUBITAK) ARDEB 3501 Grant No 120E514.  The statements made herein are solely the responsibility of the authors. 

\bibliography{references}
\bibliographystyle{aaai21}

\end{document}

%% file: stance_detection_all.tex
\pgfplotstableread{
0 0.749
1 0.695
2 0.635
3 0.579
4 0.583
}\AddHashtagAT

\pgfplotstableread{
0	0.749
1 0.633
2 0.612
3 0.585
4 0.573
}\AddSpaceAT

\pgfplotstableread{
0	0.749
1 0.627
2 0.582
3 0.568
4 0.531
}\ChangeCharacterAT

\pgfplotstableread{
0	0.749
1 0.746
2 0.694
3 0.655
4 0.626
}\NewHashtagAT

\pgfplotstableread{
0	0.749
1	0.707
2	0.729
3	0.717
4	0.704
}\RemoveHashtagAT

\pgfplotstableread{
0	0.749
1	0.710
2	0.725
3	0.710
4	0.709
}\RemoveSpaceAT

\pgfplotstableread{
0	0.749
1	0.632
2	0.699
3	0.604
4	0.614
}\ShuffleAT

\pgfplotstableread{
0	0.514
1	0.519
2	0.449
3	0.412
4	0.412
}\AddHashtagCC

\pgfplotstableread{
0	0.514
1	0.499
2	0.467
3	0.456
4	0.388
}\AddSpaceCC

\pgfplotstableread{
0	0.514
1	0.509
2	0.494
3	0.456
4	0.410
}\ChangeCharacterCC

\pgfplotstableread{
0	0.514
1	0.543
2	0.518
3	0.547
4	0.545
}\NewHashtagCC

\pgfplotstableread{
0	0.514
1	0.500
2	0.507
3	0.513
4	0.510
}\RemoveHashtagCC

\pgfplotstableread{
0	0.514
1	0.509
2	0.518
3	0.525
4	0.542
}\RemoveSpaceCC

\pgfplotstableread{
0	0.514
1	0.513
2	0.513
3	0.491
4	0.454
}\ShuffleCC

\pgfplotstableread{
0	0.608
1	0.605
2	0.574
3	0.570
4	0.563
}\AddHashtagFM

\pgfplotstableread{
0	0.608
1	0.568
2	0.546
3	0.518
4	0.493
}\AddSpaceFM

\pgfplotstableread{
0	0.608
1	0.595
2	0.575
3	0.528
4	0.479
}\ChangeCharacterFM

\pgfplotstableread{
0	0.608
1	0.610
2	0.606
3	0.603
4	0.596
}\NewHashtagFM

\pgfplotstableread{
0	0.608
1	0.583
2	0.569
3	0.559
4	0.559
}\RemoveHashtagFM

\pgfplotstableread{
0	0.608
1	0.556
2	0.566
3	0.550
4	0.555
}\RemoveSpaceFM

\pgfplotstableread{
0	0.608
1	0.565
2	0.520
3	0.507
4	0.453
}\ShuffleFM

\pgfplotstableread{
0	0.615
1	0.621
2	0.594
3	0.602
4	0.603
}\AddHashtagHC

\pgfplotstableread{
0	0.615
1	0.534
2	0.388
3	0.356
4	0.346
}\AddSpaceHC

\pgfplotstableread{
0	0.615
1	0.544
2	0.456
3	0.397
4	0.387
}\ChangeCharacterHC

\pgfplotstableread{
0	0.615
1	0.627
2	0.597
3	0.600
4	0.588
}\NewHashtagHC

\pgfplotstableread{
0	0.615
1	0.600
2	0.595
3	0.599
4	0.605
}\RemoveHashtagHC

\pgfplotstableread{
0	0.615
1	0.635
2	0.547
3	0.517
4	0.524
}\RemoveSpaceHC

\pgfplotstableread{
0	0.615
1	0.595
2	0.488
3	0.414
4	0.407
}\ShuffleHC

\pgfplotstableread{
0	0.604
1	0.623
2	0.616
3	0.599
4	0.614
}\AddHashtagLA

\pgfplotstableread{
0	0.604
1	0.527
2	0.475
3	0.434
4	0.443
}\AddSpaceLA

\pgfplotstableread{
0	0.604
1	0.516
2	0.486
3	0.432
4	0.419
}\ChangeCharacterLA

\pgfplotstableread{
0	0.604
1	0.596
2	0.628
3	0.597
4	0.597
}\NewHashtagLA

\pgfplotstableread{
0	0.604
1	0.568
2	0.568
3	0.573
4	0.570
}\RemoveHashtagLA

\pgfplotstableread{
0	0.604
1	0.610
2	0.602
3	0.589
4	0.581
}\RemoveSpaceLA

\pgfplotstableread{
0	0.604
1	0.545
2	0.534
3	0.473
4	0.439
}\ShuffleLA

\pgfplotsset{
  compat=newest,
  every minor tick={very thin, gray},
  minor tick num=4,
  enlargelimits=0.02,
   every tick label/.append style={font=\small},
  group style={
    columns=3,
    xlabels at=edge bottom,
    ylabels at=edge left},
  every axis legend/.append style={
    legend cell align=left,
  }
}

\begin{tikzpicture}

\begin{groupplot}[group style={group size= 5 by 1, horizontal sep=0.2cm},
        grid style          = {line width=.1pt,draw=gray!10},
        major grid  style={line width=.2pt,draw=gray!50},
        height              = 5cm, 
        width=4.5cm,
        xlabel              = N ,
        legend entries={Add HT, Add Space, Change Char., Add \# Sign, Remove HT, Remove Space, Shuffle } ,
        legend style={
            at={(-0.6,4.5)},
            anchor=north west,
            legend columns=-1,transpose legend,
            nodes={font=\small},
            /tikz/every even column/.append style={column sep=0.1cm}
        },
        grid,   
	    yticklabel style    = {/pgf/number format/precision=4}  ,
	    scaled y ticks      = false,
         every axis title shift=0,
      legend to name=grouplegend
        ]

 \nextgroupplot[title=Atheism,
 ylabel style={align=center}, 
ylabel={$F_1$},ylabel shift = -7 pt,ymin=0.3,ymax=0.75]   

        \addplot +[    ] table {\AddHashtagAT};\label{plots:addhashtag}
        \addplot +[mark options={scale=1,fill=orange},orange ] table {\AddSpaceAT};\label{plots:addspace}
        
        \addplot +[mark options={scale=1.5,fill=cyan},mark=triangle*,cyan ] table {\ChangeCharacterAT};\label{plots:cc}
        \addplot +[mark options={scale=1.5,fill=green},mark=triangle*,green ] table {\NewHashtagAT};\label{plots:newhashtag}
        \addplot +[mark options={scale=1.5,fill=yellow},gray ] table {\RemoveHashtagAT};\label{plots:removeht}
        \addplot +[ mark options={scale=1,fill=red},red] table {\RemoveSpaceAT};\label{plots:removespace}
        \addplot +[mark options={scale=1,fill=darkgray},mark=square*, darkgray ] table {\ShuffleAT};\label{plots:shuffle}

    \coordinate (top) at (rel axis cs:0,1.1);

\nextgroupplot[title=Hillary  Clinton ,ymin=0.3,ymax=0.75, yticklabels = \empty]        
        \addplot +[] table {\AddHashtagHC};
        
        \addplot +[mark options={scale=1,fill=orange},orange] table {\AddSpaceHC};
        
        \addplot +[mark options={scale=1.5,fill=cyan},mark=triangle*,cyan] table {\ChangeCharacterHC};
        \addplot +[mark options={scale=1.5,fill=green},mark=triangle*,green] table {\NewHashtagHC};
        \addplot +[mark options={scale=1.5,fill=yellow},gray] table {\RemoveHashtagHC};
        \addplot +[mark options={scale=1,fill=red},red] table {\RemoveSpaceHC};
        \addplot +[ mark options={scale=1,fill=darkgray},mark=square*, darkgray] table {\ShuffleHC};
        \node[mark=square, draw=blue!80, inner sep=0pt, minimum size=12pt] at (axis cs: 0,.07) {BERT};

\nextgroupplot[title=Feminism,ymin=0.3,ymax=0.75, yticklabels = \empty ]        
        \addplot +[    ] table {\AddHashtagFM};\label{plots:addhashtag}
        \addplot +[mark options={scale=1,fill=orange},orange ] table {\AddSpaceFM};\label{plots:addspace}
        
        \addplot +[mark options={scale=1.5,fill=cyan},mark=triangle*,cyan ] table {\ChangeCharacterFM};\label{plots:cc}
        \addplot +[mark options={scale=1.5,fill=green},mark=triangle*,green ] table {\NewHashtagFM};\label{plots:newhashtag}
        \addplot +[mark options={scale=1.5,fill=yellow},gray ] table {\RemoveHashtagFM};\label{plots:removeht}
        \addplot +[ mark options={scale=1,fill=red},red] table {\RemoveSpaceFM};\label{plots:removespace}
        \addplot +[mark options={scale=1,fill=darkgray},mark=square*, darkgray ] table {\ShuffleFM};
  
  \nextgroupplot[title=Climate Change ,ymin=0.3,ymax=0.75, yticklabels = \empty]        

        \addplot +[    ] table {\AddHashtagCC};\label{plots:addhashtag}
        \addplot +[mark options={scale=1,fill=orange},orange ] table {\AddSpaceCC};\label{plots:addspace}
        
        \addplot +[mark options={scale=1.5,fill=cyan},mark=triangle*,cyan ] table {\ChangeCharacterCC};\label{plots:cc}
        \addplot +[mark options={scale=1.5,fill=green},mark=triangle*,green ] table {\NewHashtagCC};\label{plots:newhashtag}
        \addplot +[mark options={scale=1.5,fill=yellow},gray ] table {\RemoveHashtagCC};\label{plots:removeht}
        \addplot +[ mark options={scale=1,fill=red},red] table {\RemoveSpaceCC};\label{plots:removespace}
        \addplot +[ mark options={scale=1,fill=darkgray},mark=square*, darkgray] table {\ShuffleCC};
    
\nextgroupplot[title=Abortion, ymin=0.3,ymax=0.75, yticklabels = \empty]        

        \addplot +[    ] table {\AddHashtagLA};\label{plots:addhashtag}
        \addplot +[mark options={scale=1,fill=orange},orange ] table {\AddSpaceLA};\label{plots:addspace}
        
        \addplot +[mark options={scale=1.5,fill=cyan},mark=triangle*,cyan ] table {\ChangeCharacterLA};\label{plots:cc}
        \addplot +[mark options={scale=1.5,fill=green},mark=triangle*,green ] table {\NewHashtagLA};\label{plots:newhashtag}
        \addplot +[mark options={scale=1.5,fill=yellow},gray ] table {\RemoveHashtagLA};\label{plots:removeht}
        \addplot +[ mark options={scale=1,fill=red},red] table {\RemoveSpaceLA};\label{plots:removespace}
        \addplot +[mark options={scale=1,fill=darkgray},mark=square*, darkgray ] table {\ShuffleLA};

    \coordinate (bot) at (rel axis cs:1,0);

\end{groupplot}
\node (a) at (16,1.5) {\rotatebox{270}{BERT}};  

    \ref{grouplegend}
\end{tikzpicture}

\pgfplotstableread{
0	0.653
1	0.672
2	0.691
3	0.702
4	0.700
}\AddHashtagATCB

\pgfplotstableread{
0	0.653
1	0.531
2	0.496
3	0.482
4	0.463
}\AddSpaceATCB

\pgfplotstableread{
0	0.653
1	0.475
2	0.496
3	0.499
4	0.481
}\ChangeCharacterATCB

\pgfplotstableread{
0	0.653
1	0.663
2	0.664
3	0.673
4	0.695
}\NewHashtagATCB

\pgfplotstableread{
0	0.653
1	0.669
2	0.663
3	0.680
4	0.671
}\RemoveHashtagATCB

\pgfplotstableread{
0	0.653
1	0.596
2	0.617
3	0.586
4	0.594
}\RemoveSpaceATCB

\pgfplotstableread{
0	0.653
1	0.514
2	0.516
3	0.546
4	0.523
}\ShuffleATCB

\pgfplotstableread{
0	0.489
1	0.513
2	0.408
3	0.421
4	0.417
}\AddHashtagCCCB

\pgfplotstableread{
0	0.489
1	0.501
2	0.479
3	0.447
4	0.432
}\AddSpaceCCCB

\pgfplotstableread{
0	0.489
1	0.457
2	0.458
3	0.448
4	0.437
}\ChangeCharacterCCCB

\pgfplotstableread{
0	0.489
1	0.515
2	0.505
3	0.483
4	0.487
}\NewHashtagCCCB

\pgfplotstableread{
0	0.489
1	0.491
2	0.501
3	0.508
4	0.502
}\RemoveHashtagCCCB

\pgfplotstableread{
0	0.489
1	0.518
2	0.514
3	0.507
4	0.523
}\RemoveSpaceCCCB

\pgfplotstableread{
0	0.489
1	0.462
2	0.460
3	0.443
4	0.429
}\ShuffleCCCB

\pgfplotstableread{
0	0.517
1	0.562
2	0.513
3	0.501
4	0.504
}\AddHashtagFMCB

\pgfplotstableread{
0	0.517
1	0.515
2	0.494
3	0.432
4	0.428
}\AddSpaceFMCB

\pgfplotstableread{
0	0.517
1	0.492
2	0.497
3	0.516
4	0.499
}\ChangeCharacterFMCB

\pgfplotstableread{
0	0.517
1	0.524
2	0.551
3	0.549
4	0.545
}\NewHashtagFMCB

\pgfplotstableread{
0	0.517
1	0.506
2	0.449
3	0.461
4	0.445
}\RemoveHashtagFMCB

\pgfplotstableread{
0	0.517
1	0.496
2	0.485
3	0.507
4	0.500
}\RemoveSpaceFMCB

\pgfplotstableread{
0	0.517
1	0.487
2	0.469
3	0.460
4	0.452
}\ShuffleFMCB

\pgfplotstableread{
0	0.432
1	0.441
2	0.400
3	0.379
4	0.345
}\AddHashtagHCCB

\pgfplotstableread{
0	0.432
1	0.434
2	0.388
3	0.311
4	0.301
}\AddSpaceHCCB

\pgfplotstableread{
0	0.432
1	0.409
2	0.371
3	0.355
4	0.322
}\ChangeCharacterHCCB

\pgfplotstableread{
0	0.432
1	0.433
2	0.408
3	0.409
4	0.413
}\NewHashtagHCCB

\pgfplotstableread{
0	0.432
1	0.473
2	0.478
3	0.482
4	0.489
}\RemoveHashtagHCCB

\pgfplotstableread{
0	0.432
1	0.467
2	0.451
3	0.415
4	0.395
}\RemoveSpaceHCCB

\pgfplotstableread{
0	0.432
1	0.443
2	0.425
3	0.388
4	0.375
}\ShuffleHCCB

\pgfplotstableread{
0	0.370
1	0.387
2	0.397
3	0.397
4	0.389
}\AddHashtagLACB

\pgfplotstableread{
0	0.370
1	0.369
2	0.346
3	0.340
4	0.309
}\AddSpaceLACB

\pgfplotstableread{
0	0.370
1	0.344
2	0.322
3	0.311
4	0.294
}\ChangeCharacterLACB

\pgfplotstableread{
0	0.370
1	0.392
2	0.388
3	0.399
4	0.394
}\NewHashtagLACB

\pgfplotstableread{
0	0.370
1	0.381
2	0.374
3	0.372
4	0.367
}\RemoveHashtagLACB

\pgfplotstableread{
0	0.370
1	0.372
2	0.364
3	0.370
4	0.365
}\RemoveSpaceLACB

\pgfplotstableread{
0	0.370
1	0.374
2	0.352
3	0.354
4	0.324
}\ShuffleLACB

\pgfplotsset{
  compat=newest,
  every minor tick={very thin, gray},
  minor tick num=4,
  enlargelimits=0.02,
   every tick label/.append style={font=\small},
  group style={
    columns=3,
    xlabels at=edge bottom,
    ylabels at=edge left},
  every axis legend/.append style={
    legend cell align=left,
  }
}

%% file: roberta.tex
\pgfplotstableread{
0	0.619
1	0.641
2	0.651
3	0.667
4	0.677
}\AddHashtagAT

\pgfplotstableread{
0	0.619
1	0.554
2	0.550
3	0.530
4	0.521
}\AddSpaceAT

\pgfplotstableread{
0	0.619
1	0.559
2	0.524
3	0.476
4	0.471
}\ChangeCharacterAT

\pgfplotstableread{
0	0.619
1	0.627
2	0.647
3	0.652
4	0.663
}\NewHashtagAT

\pgfplotstableread{
0	0.619
1	0.603
2	0.588
3	0.593
4	0.587
}\RemoveHashtagAT

\pgfplotstableread{
0	0.619
1	0.608
2	0.638
3	0.638
4	0.632
}\RemoveSpaceAT

\pgfplotstableread{
0	0.619
1	0.586
2	0.583
3	0.565
4	0.563
}\ShuffleAT

\pgfplotstableread{
0	0.528
1	0.548
2	0.393
3	0.393
4	0.393
}\AddHashtagCC

\pgfplotstableread{
0	0.528
1	0.456
2	0.413
3	0.360
4	0.325
}\AddSpaceCC

\pgfplotstableread{
0	0.528
1	0.489
2	0.450
3	0.434
4	0.406
}\ChangeCharacterCC

\pgfplotstableread{
0	0.528
1	0.514
2	0.531
3	0.534
4	0.527
}\NewHashtagCC

\pgfplotstableread{
0	0.528
1	0.523
2	0.503
3	0.493
4	0.501
}\RemoveHashtagCC

\pgfplotstableread{
0	0.528
1	0.524
2	0.524
3	0.510
4	0.499
}\RemoveSpaceCC

\pgfplotstableread{
0	0.528
1	0.469
2	0.444
3	0.395
4	0.385
}\ShuffleCC

\pgfplotstableread{
0	0.588
1	0.569
2	0.540
3	0.518
4	0.563
}\AddHashtagFM

\pgfplotstableread{
0	0.588
1	0.559
2	0.537
3	0.500
4	0.515
}\AddSpaceFM

\pgfplotstableread{
0	0.588
1	0.558
2	0.559
3	0.535
4	0.528
}\ChangeCharacterFM

\pgfplotstableread{
0	0.588
1	0.534
2	0.573
3	0.580
4	0.588
}\NewHashtagFM

\pgfplotstableread{
0	0.588
1	0.550
2	0.535
3	0.533
4	0.534
}\RemoveHashtagFM

\pgfplotstableread{
0	0.588
1	0.570
2	0.556
3	0.571
4	0.588
}\RemoveSpaceFM

\pgfplotstableread{
0	0.588
1	0.545
2	0.503
3	0.536
4	0.504
}\ShuffleFM

\pgfplotstableread{
0	0.574
1	0.577
2	0.560
3	0.575
4	0.580
}\AddHashtagHC

\pgfplotstableread{
0	0.574
1	0.560
2	0.463
3	0.413
4	0.405
}\AddSpaceHC

\pgfplotstableread{
0	0.574
1	0.514
2	0.484
3	0.459
4	0.454
}\ChangeCharacterHC

\pgfplotstableread{
0	0.574
1	0.566
2	0.557
3	0.567
4	0.556
}\NewHashtagHC

\pgfplotstableread{
0	0.574
1	0.566
2	0.549
3	0.538
4	0.543
}\RemoveHashtagHC

\pgfplotstableread{
0	0.574
1	0.568
2	0.584
3	0.557
4	0.551
}\RemoveSpaceHC

\pgfplotstableread{
0	0.574
1	0.526
2	0.490
3	0.462
4	0.424
}\ShuffleHC

\pgfplotstableread{
0	0.514
1	0.527
2	0.513
3	0.523
4	0.528
}\AddHashtagLA

\pgfplotstableread{
0	0.514
1	0.505
2	0.492
3	0.445
4	0.373
}\AddSpaceLA

\pgfplotstableread{
0	0.514
1	0.497
2	0.482
3	0.437
4	0.397
}\ChangeCharacterLA

\pgfplotstableread{
0	0.514
1	0.529
2	0.545
3	0.558
4	0.555
}\NewHashtagLA

\pgfplotstableread{
0	0.514
1	0.498
2	0.489
3	0.490
4	0.499
}\RemoveHashtagLA

\pgfplotstableread{
0	0.514
1	0.517
2	0.505
3	0.512
4	0.499
}\RemoveSpaceLA

\pgfplotstableread{
0	0.514
1	0.503
2	0.471
3	0.428
4	0.383
}\ShuffleLA

\pgfplotsset{
  compat=newest,
  every minor tick={very thin, gray},
  minor tick num=4,
  enlargelimits=0.02,
   every tick label/.append style={font=\small},
  group style={
    columns=3,
    xlabels at=edge bottom,
    ylabels at=edge left},
  every axis legend/.append style={
    legend cell align=left,
  }
}

\begin{tikzpicture}

\begin{groupplot}[group style={group size= 5 by 1, horizontal sep=0.2cm},
        grid style          = {line width=.1pt,draw=gray!10},
        major grid  style={line width=.2pt,draw=gray!50},
        height              = 5cm, 
        width=4.5cm,
        xlabel              = N ,
        grid,   
	    yticklabel style    = {/pgf/number format/precision=4}  ,
	    scaled y ticks      = false,
         every axis title shift=0,
      legend to name=grouplegend
        ]
 \nextgroupplot[title=Atheism,
 ylabel style={align=center}, 
ylabel={$F_1$},ylabel shift = -7 pt,ymin=0.3,ymax=0.75]   

        \addplot +[    ] table {\AddHashtagAT};\label{plots:addhashtag}
        \addplot +[mark options={scale=1,fill=orange},orange ] table {\AddSpaceAT};\label{plots:addspace}
        
        \addplot +[mark options={scale=1.5,fill=cyan},mark=triangle*,cyan ] table {\ChangeCharacterAT};\label{plots:cc}
        \addplot +[mark options={scale=1.5,fill=green},mark=triangle*,green ] table {\NewHashtagAT};\label{plots:newhashtag}
        \addplot +[mark options={scale=1.5,fill=yellow},gray ] table {\RemoveHashtagAT};\label{plots:removeht}
        \addplot +[ mark options={scale=1,fill=red},red] table {\RemoveSpaceAT};\label{plots:removespace}
        \addplot +[mark options={scale=1,fill=darkgray},mark=square*, darkgray ] table {\ShuffleAT};\label{plots:shuffle}

    \coordinate (top) at (rel axis cs:0,1.1);

\nextgroupplot[title=Hillary  Clinton ,ymin=0.3,ymax=0.75, yticklabels = \empty]        
        \addplot +[] table {\AddHashtagHC};
        
        \addplot +[mark options={scale=1,fill=orange},orange] table {\AddSpaceHC};
        
        \addplot +[mark options={scale=1.5,fill=cyan},mark=triangle*,cyan] table {\ChangeCharacterHC};
        \addplot +[mark options={scale=1.5,fill=green},mark=triangle*,green] table {\NewHashtagHC};
        \addplot +[mark options={scale=1.5,fill=yellow},gray] table {\RemoveHashtagHC};
        \addplot +[mark options={scale=1,fill=red},red] table {\RemoveSpaceHC};
        \addplot +[ mark options={scale=1,fill=darkgray},mark=square*, darkgray] table {\ShuffleHC};
        \node[mark=square, draw=blue!80, inner sep=0pt, minimum size=12pt] at (axis cs: 0,.07) {BERT};

\nextgroupplot[title=Feminism,ymin=0.3,ymax=0.75, yticklabels = \empty ]        
        \addplot +[    ] table {\AddHashtagFM};\label{plots:addhashtag}
        \addplot +[mark options={scale=1,fill=orange},orange ] table {\AddSpaceFM};\label{plots:addspace}
        
        \addplot +[mark options={scale=1.5,fill=cyan},mark=triangle*,cyan ] table {\ChangeCharacterFM};\label{plots:cc}
        \addplot +[mark options={scale=1.5,fill=green},mark=triangle*,green ] table {\NewHashtagFM};\label{plots:newhashtag}
        \addplot +[mark options={scale=1.5,fill=yellow},gray ] table {\RemoveHashtagFM};\label{plots:removeht}
        \addplot +[ mark options={scale=1,fill=red},red] table {\RemoveSpaceFM};\label{plots:removespace}
        \addplot +[mark options={scale=1,fill=darkgray},mark=square*, darkgray ] table {\ShuffleFM};
  
  \nextgroupplot[title=Climate Change ,ymin=0.3,ymax=0.75, yticklabels = \empty]        
        \addplot +[    ] table {\AddHashtagCC};\label{plots:addhashtag}
        \addplot +[mark options={scale=1,fill=orange},orange ] table {\AddSpaceCC};\label{plots:addspace}
        
        \addplot +[mark options={scale=1.5,fill=cyan},mark=triangle*,cyan ] table {\ChangeCharacterCC};\label{plots:cc}
        \addplot +[mark options={scale=1.5,fill=green},mark=triangle*,green ] table {\NewHashtagCC};\label{plots:newhashtag}
        \addplot +[mark options={scale=1.5,fill=yellow},gray ] table {\RemoveHashtagCC};\label{plots:removeht}
        \addplot +[ mark options={scale=1,fill=red},red] table {\RemoveSpaceCC};\label{plots:removespace}
        \addplot +[ mark options={scale=1,fill=darkgray},mark=square*, darkgray] table {\ShuffleCC};
    
\nextgroupplot[title=Abortion, ymin=0.3,ymax=0.75, yticklabels = \empty]        

        \addplot +[    ] table {\AddHashtagLA};\label{plots:addhashtag}
        \addplot +[mark options={scale=1,fill=orange},orange ] table {\AddSpaceLA};\label{plots:addspace}
        
        \addplot +[mark options={scale=1.5,fill=cyan},mark=triangle*,cyan ] table {\ChangeCharacterLA};\label{plots:cc}
        \addplot +[mark options={scale=1.5,fill=green},mark=triangle*,green ] table {\NewHashtagLA};\label{plots:newhashtag}
        \addplot +[mark options={scale=1.5,fill=yellow},gray ] table {\RemoveHashtagLA};\label{plots:removeht}
        \addplot +[ mark options={scale=1,fill=red},red] table {\RemoveSpaceLA};\label{plots:removespace}
        \addplot +[mark options={scale=1,fill=darkgray},mark=square*, darkgray ] table {\ShuffleLA};

    \coordinate (bot) at (rel axis cs:1,0);

\end{groupplot}

\node (a) at (16,1.5) {\rotatebox{270}{Twitter-RoBERTa}};  

    \ref{grouplegend}
\end{tikzpicture}

\pgfplotstableread{
0	0.653
1	0.672
2	0.691
3	0.702
4	0.700
}\AddHashtagATCB

\pgfplotstableread{
0	0.653
1	0.531
2	0.496
3	0.482
4	0.463
}\AddSpaceATCB

\pgfplotstableread{
0	0.653
1	0.475
2	0.496
3	0.499
4	0.481
}\ChangeCharacterATCB

\pgfplotstableread{
0	0.653
1	0.663
2	0.664
3	0.673
4	0.695
}\NewHashtagATCB

\pgfplotstableread{
0	0.653
1	0.669
2	0.663
3	0.680
4	0.671
}\RemoveHashtagATCB

\pgfplotstableread{
0	0.653
1	0.596
2	0.617
3	0.586
4	0.594
}\RemoveSpaceATCB

\pgfplotstableread{
0	0.653
1	0.514
2	0.516
3	0.546
4	0.523
}\ShuffleATCB

\pgfplotstableread{
0	0.489
1	0.513
2	0.408
3	0.421
4	0.417
}\AddHashtagCCCB

\pgfplotstableread{
0	0.489
1	0.501
2	0.479
3	0.447
4	0.432
}\AddSpaceCCCB

\pgfplotstableread{
0	0.489
1	0.457
2	0.458
3	0.448
4	0.437
}\ChangeCharacterCCCB

\pgfplotstableread{
0	0.489
1	0.515
2	0.505
3	0.483
4	0.487
}\NewHashtagCCCB

\pgfplotstableread{
0	0.489
1	0.491
2	0.501
3	0.508
4	0.502
}\RemoveHashtagCCCB

\pgfplotstableread{
0	0.489
1	0.518
2	0.514
3	0.507
4	0.523
}\RemoveSpaceCCCB

\pgfplotstableread{
0	0.489
1	0.462
2	0.460
3	0.443
4	0.429
}\ShuffleCCCB

\pgfplotstableread{
0	0.517
1	0.562
2	0.513
3	0.501
4	0.504
}\AddHashtagFMCB

\pgfplotstableread{
0	0.517
1	0.515
2	0.494
3	0.432
4	0.428
}\AddSpaceFMCB

\pgfplotstableread{
0	0.517
1	0.492
2	0.497
3	0.516
4	0.499
}\ChangeCharacterFMCB

\pgfplotstableread{
0	0.517
1	0.524
2	0.551
3	0.549
4	0.545
}\NewHashtagFMCB

\pgfplotstableread{
0	0.517
1	0.506
2	0.449
3	0.461
4	0.445
}\RemoveHashtagFMCB

\pgfplotstableread{
0	0.517
1	0.496
2	0.485
3	0.507
4	0.500
}\RemoveSpaceFMCB

\pgfplotstableread{
0	0.517
1	0.487
2	0.469
3	0.460
4	0.452
}\ShuffleFMCB

\pgfplotstableread{
0	0.432
1	0.441
2	0.400
3	0.379
4	0.345
}\AddHashtagHCCB

\pgfplotstableread{
0	0.432
1	0.434
2	0.388
3	0.311
4	0.301
}\AddSpaceHCCB

\pgfplotstableread{
0	0.432
1	0.409
2	0.371
3	0.355
4	0.322
}\ChangeCharacterHCCB

\pgfplotstableread{
0	0.432
1	0.433
2	0.408
3	0.409
4	0.413
}\NewHashtagHCCB

\pgfplotstableread{
0	0.432
1	0.473
2	0.478
3	0.482
4	0.489
}\RemoveHashtagHCCB

\pgfplotstableread{
0	0.432
1	0.467
2	0.451
3	0.415
4	0.395
}\RemoveSpaceHCCB

\pgfplotstableread{
0	0.432
1	0.443
2	0.425
3	0.388
4	0.375
}\ShuffleHCCB

\pgfplotstableread{
0	0.370
1	0.387
2	0.397
3	0.397
4	0.389
}\AddHashtagLACB

\pgfplotstableread{
0	0.370
1	0.369
2	0.346
3	0.340
4	0.309
}\AddSpaceLACB

\pgfplotstableread{
0	0.370
1	0.344
2	0.322
3	0.311
4	0.294
}\ChangeCharacterLACB

\pgfplotstableread{
0	0.370
1	0.392
2	0.388
3	0.399
4	0.394
}\NewHashtagLACB

\pgfplotstableread{
0	0.370
1	0.381
2	0.374
3	0.372
4	0.367
}\RemoveHashtagLACB

\pgfplotstableread{
0	0.370
1	0.372
2	0.364
3	0.370
4	0.365
}\RemoveSpaceLACB

\pgfplotstableread{
0	0.370
1	0.374
2	0.352
3	0.354
4	0.324
}\ShuffleLACB

\pgfplotsset{
  compat=newest,
  every minor tick={very thin, gray},
  minor tick num=4,
  enlargelimits=0.02,
   every tick label/.append style={font=\small},
  group style={
    columns=3,
    xlabels at=edge bottom,
    ylabels at=edge left},
  every axis legend/.append style={
    legend cell align=left,
  }
}

%% file: readability_and_semantic.tex
\pgfplotstableread{
1	1
2	1
3	1
4	1
}\AddHashtagAT

\pgfplotstableread{
1	1
2	1
3	0.8
4	1
}\AddSpaceAT

\pgfplotstableread{
1	1
2	1
3	1
4	1
}\ChangeCharacterAT

\pgfplotstableread{
1	1
2	1
3	1
4	1
}\NewHashtagAT

\pgfplotstableread{
1	1
2	1
3	1
4	1
}\RemoveHashtagAT

\pgfplotstableread{
1	1
2	1
3	1
4	1
}\RemoveSpaceAT

\pgfplotstableread{
1	1
2	0.6
3	1
4	0.6
}\ShuffleAT

\pgfplotstableread{
1	1
2	1
3	1
4	1
}\AddHashtagHC

\pgfplotstableread{
1	1
2	1
3	0.8
4	1
}\AddSpaceHC

\pgfplotstableread{
1	1
2	1
3	1
4	1
}\ChangeCharacterHC

\pgfplotstableread{
1	1
2	1
3	1
4	1
}\NewHashtagHC

\pgfplotstableread{
1	0.6
2	0.8
3	0.6
4	1
}\RemoveHashtagHC

\pgfplotstableread{
1	1
2	1
3	1
4	1
}\RemoveSpaceHC

\pgfplotstableread{
1	1
2	1
3	1
4	1
}\ShuffleHC

\pgfplotsset{
  compat=newest,
  every minor tick={very thin, gray},
  minor tick num=4,
  enlargelimits=0.02,
   every tick label/.append style={font=\small},
  group style={
    columns=3,
    xlabels at=edge bottom,
    ylabels at=edge left},
  every axis legend/.append style={
    legend cell align=left,
  }
}

\begin{tikzpicture}

\begin{groupplot}[group style={group size= 2 by 1, horizontal sep=2cm},
        grid style          = {line width=.1pt,draw=gray!10},
        major grid  style={line width=.2pt,draw=gray!50},
        height              = 5.5cm, 
        width=6cm,
        legend entries={Add HT, Add Space, Change Char., Add \# Sign, Remove HT, Remove Space, Shuffle } ,
        legend style={
            at={(-4,4)},
            anchor=north west,
            legend columns=1,transpose legend,
            nodes={font=\small},
            /tikz/every even column/.append style={column sep=0.1cm}
        },
        grid,   
	    yticklabel style    = {/pgf/number format/precision=4}  ,
	    scaled y ticks      = true,
         every axis title shift=0,
      legend to name=grouplegend
        ]

 \nextgroupplot[title=Readability, font=\small, 
 ylabel style={align=center},  xlabel={N},
ylabel={Ratio of Readable Tweets},ylabel shift = -4pt, ymin=0,ymax=1.05]   

        \addplot +[    ] table {\AddHashtagAT};\label{plots:addhashtag}
        \addplot +[mark options={scale=1,fill=orange},orange ] table {\AddSpaceAT};\label{plots:addspace}
        
        \addplot +[mark options={scale=1.5,fill=cyan},mark=triangle*,cyan ] table {\ChangeCharacterAT};\label{plots:cc}
        \addplot +[mark options={scale=1.5,fill=green},mark=triangle*,green ] table {\NewHashtagAT};\label{plots:newhashtag}
        \addplot +[mark options={scale=1.5,fill=yellow},gray ] table {\RemoveHashtagAT};\label{plots:removeht}
        \addplot +[ mark options={scale=1,fill=red},red] table {\RemoveSpaceAT};\label{plots:removespace}
        \addplot +[mark options={scale=1,fill=darkgray},mark=square*, darkgray ] table {\ShuffleAT};\label{plots:shuffle}

    \coordinate (top) at (rel axis cs:0,1.1);

\nextgroupplot[title=Semantic ,ymin=0,ymax=1.05, ylabel = {Ratio of No-Semantic-Change},  xlabel={N}, font=\small, ylabel shift = -5.5 pt]        
        \addplot +[] table {\AddHashtagHC};
        
        \addplot +[mark options={scale=1,fill=orange},orange] table {\AddSpaceHC};
        
        \addplot +[mark options={scale=1.5,fill=cyan},mark=triangle*,cyan] table {\ChangeCharacterHC};
        \addplot +[mark options={scale=1.5,fill=green},mark=triangle*,green] table {\NewHashtagHC};
        \addplot +[mark options={scale=1.5,fill=yellow},gray] table {\RemoveHashtagHC};
        \addplot +[mark options={scale=1,fill=red},red] table {\RemoveSpaceHC};
        \addplot +[ mark options={scale=1,fill=darkgray},mark=square*, darkgray] table {\ShuffleHC};
        \node[mark=square, draw=blue!80, inner sep=0pt, minimum size=12pt] at (axis cs: 0,.07) {BERT};

    \coordinate (bot) at (rel axis cs:1,0);

\end{groupplot}

    \ref{grouplegend}
\end{tikzpicture}

%% file: geotagging_gcn.tex
\pgfplotstableread{
0 540
1 538
2 546
3 545
4 549
}\AddSpaceGCN

\pgfplotstableread{
0 540
1 548
2 539
3 539
4 542
}\ChangeCharGCN

\pgfplotstableread{
0 540
1 541
2 543
3 532
4 545
}\NewHashtagGCN

\pgfplotstableread{
0 540
1 540
2 540
3 540
4 540
}\RemoveHashtagGCN

\pgfplotstableread{
0 540
1 544
2 557
3 558
4 561
}\RemoveSpaceGCN

\pgfplotstableread{
0 540
1 538
2 529
3 535
4 529
}\ShuffleGCN

\pgfplotstableread{
0 553
1 591
2 599
3 636
4 633
}\AddSpaceMLP

\pgfplotstableread{
0 553
1 591
2 611
3 631
4 637
}\ChangeCharMLP

\pgfplotstableread{
0 553
1 580
2 556
3 576
4 573
}\NewHashtagMLP

\pgfplotstableread{
0 553
1 554
2 554
2 554
4 554
}\RemoveHashtagMLP

\pgfplotstableread{
0 553
1 553
2 575
3 581
4 582
}\RemoveSpaceMLP

\pgfplotstableread{
0 553
1 558
2 558
3 576
4 578
}\ShuffleMLP

\pgfplotsset{
  compat=newest,
  every minor tick={very thin, gray},
  minor tick num=4,
  enlargelimits=0.02,
  every tick label/.append style={font=\small},
  group style={
    columns=2,
    xlabels at=edge bottom,
    ylabels at=edge left},
  every axis legend/.append style={
    legend cell align=left,
  }
}

\begin{tikzpicture}

\begin{groupplot}[group style={group size= 2 by 1,horizontal sep=2cm},
        grid style          = {line width=.1pt,draw=gray!10},
        major grid  style={line width=.2pt,draw=gray!50},
        height              = 5.5cm, 
        width=6cm,
        xlabel              = N ,
        legend entries={Add Space, Change Char., Add \# Sign, Remove HT, Remove Space, Shuffle } ,
        legend style={
            at={(-4.2,4)},
            anchor=north west,
            legend columns=1,transpose legend,
            nodes={font=\small},
            /tikz/every even column/.append style={column sep=0.1cm}
        },
        grid,   
	    yticklabel style    = {/pgf/number format/precision=4}  ,
	    scaled y ticks      = false,
         every axis title shift=0,
      legend to name=grouplegend,
        ]

 \nextgroupplot[title=GCN,title style={at={(0,1)},anchor=north west},ylabel={Mean Error},ymax=650,ymin=525]        
        \addplot +[mark options={scale=1.5,fill=orange},orange ] table {\AddSpaceGCN};
        \addplot +[mark options={scale=1.5,fill=cyan},mark=triangle*,cyan ] table {\ChangeCharGCN};
        \addplot +[mark options={scale=1.5,fill=green},mark=triangle*,green  ] table {\NewHashtagGCN};
        \addplot +[ mark options={scale=2,fill=yellow},mark=diamond*,gray ] table {\RemoveHashtagGCN};
        \addplot +[ mark options={scale=1.5,fill=red},red] table {\RemoveSpaceGCN};
        \addplot +[ mark options={scale=1,fill=darkgray},mark=square*, darkgray ] table {\ShuffleGCN};
  
\hfill
 \nextgroupplot[title=MLP-TXT+NET,title style={at={(0,1)},anchor=north west}, ymax=650,ymin=525, ylabel={Mean Error}]        
        \addplot +[  mark options={scale=1.5,fill=orange},orange  ] table {\AddSpaceMLP};
        \addplot +[mark options={scale=1.5,fill=cyan},mark=triangle*,cyan ] table {\ChangeCharMLP};
        
        \addplot +[ mark options={scale=1.5,fill=green},mark=triangle*,green ] table {\NewHashtagMLP};
        \addplot +[mark options={scale=2,fill=yellow},mark=diamond*,gray] table {\RemoveHashtagMLP};
        \addplot +[mark options={scale=1.5,fill=red},red ] table {\RemoveSpaceMLP};
         \addplot +[mark options={scale=1,fill=darkgray},mark=square*, darkgray ] table {\ShuffleMLP};


\end{groupplot}

    \ref{grouplegend}
\end{tikzpicture}

%% file: additional_tweets.tex
\pgfplotstableread{
0 540
10\% 538
20\% 539
30\% 535
40\% 536
50\% 538
}\MentionCityGCNa

\pgfplotstableread{
0 553
10\% 572
20\% 574
30\% 586
40\% 582
50\% 585
}\MentionCityMLPa

\pgfplotstableread{
0 540
10\% 768
20\% 884
30\% 972
40\% 1021
50\% 1060
}\MentionUserGCNa

\pgfplotstableread{
0 553
10\% 730
20\% 761
30\% 809
40\% 826
50\% 825
}\MentionUserMLPa

\begin{tikzpicture}
\begin{axis}[
        grid style          = {line width=.1pt,draw=gray!10},
        major grid style={line width=.2pt,draw=gray!50},
        height              = 7cm, 
        width=7.5cm ,
        ylabel              = Mean Error,
        xlabel              = Tweet Increment Ratio per Profile ,
        grid,
        ymin                = 0,
	    yticklabel style    = {/pgf/number format/precision=4}  ,
	    scaled y ticks      = false                             ,
	    xmin                = 0                               ,
	    xmax                = 50\%                                 ,
	    symbolic x coords={ 0, 10\%,20\%,30\%,40\%,50\% },
	    legend style={at={(0,0)},anchor=south west,font=\small}   ,
        every node near coord/.style=above left                 
        ]

    
        \addplot+ [mark options={solid,scale=2,fill=cyan},mark=square,cyan,very thick] table {\MentionCityGCNa};
        
        \addplot +[mark options={solid,scale=2,fill=cyan},dashed,mark=square,gray,very thick] table {\MentionUserGCNa};
        
        \addplot +[mark options={solid,scale=2,fill=gray},mark=diamond,cyan,very thick] table {\MentionCityMLPa};

        \addplot +[mark options={solid,scale=2,fill=gray},dashed,mark=diamond,gray,very thick] table {\MentionUserMLPa};
     
        \legend{GCN w/ MentionCity, GCN w/ MentionUser,MLP-TXT+NET w/ MentionCity, MLP-TXT+NET w/ MentionUser}
      
\end{axis}
\end{tikzpicture}